\documentclass[pdflatex]{sn-jnl}
\jyear{2026}

\usepackage{graphicx} \graphicspath{{figures/}}
\usepackage{acronym}
\usepackage{enumitem}
\usepackage{xspace}
\usepackage[capitalise,noabbrev,nameinlink]{cleveref}
\usepackage{booktabs,tabularx,colortbl,multirow,multicol,array,makecell,tabularray}
\usepackage{pifont}

\makeatletter
\DeclareRobustCommand\onedot{\futurelet\@let@token\@onedot}
\def\@onedot{\ifx\@let@token.\else.\null\fi\xspace}
\def\eg{\emph{e.g}\onedot} 

\def\ie{\emph{i.e}\onedot}

\makeatother

\makeatletter
\renewcommand{\paragraph}{%
  \@startsection{paragraph}{4}%
  {\z@}{0ex \@plus 0ex \@minus 0ex}{-1em}%
  {\hskip\parindent\normalfont\normalsize\bfseries}%
}
\makeatother

\crefname{algorithm}{Alg.}{Algs.}
\Crefname{algocf}{Algorithm}{Algorithms}
\crefname{section}{Sec.}{Secs.}
\Crefname{section}{Section}{Sections}
\crefname{table}{Tab.}{Tabs.}
\Crefname{table}{Table}{Tables}
\crefname{figure}{Fig.}{Figs.}
\Crefname{figure}{Figure}{Figures}
\crefname{equation}{Eq.}{Eqs.}
\Crefname{equation}{Equation}{Equations}
\crefname{appendix}{Supp.}{Supps.}
\Crefname{appendix}{Appendix}{Appendices}

\definecolor{gblue}{HTML}{4285F4}
\definecolor{gred}{HTML}{DB4437}
\definecolor{ggreen}{HTML}{0F9D58}

\usepackage[mathlines]{lineno}
\usepackage{bm}
\usepackage{bbm}

\acrodef{llm}[LLM]{Large Language Model}
\acrodef{ai}[AI]{Artificial Intelligence}
\acrodef{md}[MD]{multiple-demand}
\acrodef{pg}[PG]{precentral gyrus}
\acrodef{ppc}[PPC]{posterior parietal cortex}
\acrodef{mfg}[MFG]{middle frontal gyrus}
\acrodef{ifg}[IFG]{inferior frontal gyrus}
\acrodef{mefg}[MeFG]{medial frontal gyrus}
\acrodef{bg}[BG]{Basal Ganglia}
\acrodef{mse}[MSE]{mean-square-error}
\acrodef{ce}[CE]{cross-entropy}
\acrodef{cot}[CoT]{Chain-of-Thought}
\acrodef{mni}[MNI]{Montreal Neurological Institute}
\acrodef{glm}[GLM]{General Linear Model}
\acrodef{ffn}[FFN]{Feed-Forward Network}
\acrodef{mad}[m.a.d.]{median absolute deviation}
\acrodef{nari}[NARI]{Neural Activation guided Representation Intervention}
\acrodef{narf}[NARF]{Neural Activation guided Representation Fine-tuning}
\acrodef{fmri}[fMRI]{functional magnetic resonance imaging}

\acrodef{pca}[PCA]{principal component analysis}
\acrodef{lora}[LoRA]{Low-Rank Adaptation}
\acrodef{hrf}[HRF]{hemodynamic response function}

\makeatletter
\def\ps@headings{%
    \def\@oddfoot{\hfill}%
    \let\@evenfoot\@oddfoot%
    \def\@evenhead{\hbox to \hsize{\headerfont\thepage\qquad\rightmark\hfill}}%
    \def\@oddhead{\hbox to \hsize{\headerfont\hfill\leftmark\qquad\thepage}}%
    \let\@mkboth\markboth%
}
\def\ps@titlepage{%
    \let\@oddhead\@empty\let\@evenhead\@empty%
    \def\@oddfoot{\vbox to 18pt{\vfill\reset@font\rmfamily\hfil\thepage\hfil}}%
    \def\@evenfoot{}%
}
\makeatother
\begin{document}


\title[Brain-guided language models for robust reasoning]{Beyond representational alignment with brain-guided language models for robust reasoning}

\author[1,3]{\fnm{Mingqing} \sur{Xiao}}\email{mingqing\_xiao@pku.edu.cn}
\author*[2]{\fnm{Kai} \sur{Du}}\email{kai\_du@tsinghua.edu.cn}
\author*[1]{\fnm{Zhouchen} \sur{Lin}}\email{zlin@pku.edu.cn}
\affil[1]{\orgdiv{State Key Lab of General AI, School of Intelligence Science and Technology}, \orgname{Peking University}}
\affil[2]{\orgdiv{Department of Psychological and Cognitive Sciences}, \orgname{Tsinghua University}}
\affil[3]{\orgname{Microsoft Research Asia}}

\abstract{
The correspondence between large language models (LLMs) and the neural mechanisms underlying human higher-order cognition remains insufficiently characterized. Given that language and reasoning in the human brain appear dissociable, an open question is whether LLMs align with neural signals from reasoning-related regions and whether such signals can improve them. Here, focusing on deductive reasoning, we show that LLM internal representations are not only partially aligned with task-fMRI activity but can also be directly enhanced by these signals. Using a neural-predictivity metric, we find that LLMs explain a substantial fraction of the explainable variance in reasoning-related regions at the aggregate level, whereas predictivity within specific reasoning types is lower, indicating both alignment and divergence. Building on this, we propose a brain-guided framework: we steer model representations along directions induced by the joint structure of model and brain representations, applying intervention at inference and fine-tuning during training. We demonstrate that task-evoked brain signals can directly enhance LLM reasoning, yielding gains orthogonal to language-only supervision across 10 LLMs (1.5B-72B), with transfer across reasoning types and up to 13\% absolute accuracy gain. Our results advance LLM-brain correspondences from correlation to guidance, establishing a brain-signal-driven pathway toward more robust and cognitively aligned AI.
}


\maketitle

\section{Introduction}

The emergence of \acfp{llm} has rapidly transformed the landscape of artificial intelligence and even begun to inform cognitive neuroscience. These models now demonstrate increasingly human-like performance on higher-order cognitive tasks beyond pure language processing (for example, in logical reasoning) ~\cite{wei2022emergent,wei2022chain,webb2023emergent,saparov2023testing}. Deductive reasoning---the ability to draw logically necessary conclusions from given premises---is widely regarded as a core human cognitive skill for logical thought~\cite{rips1994psychology,prado2011brain}. Crucially, it enables the manipulation of abstract rules and symbolic relationships that transcend natural language, forming the foundation of formal disciplines like logic and mathematics~\cite{tarski1994introduction}. Evidence from cognitive neuroscience suggests that this cognitive process elicits characteristic patterns of brain activity in specific regions measurable with neuroimaging~\cite{monti2009boundaries,prado2011brain}, and the involved brain regions are largely dissociable from language areas~\cite{monti2009boundaries,prado2011brain,fedorenko2024language}. Interestingly, the capability of deductive reasoning has also been observed, to some degree, in \ac{llm} behaviour under certain conditions~\cite{saparov2023testing,seals2024evaluating}, although their reasoning abilities remain imperfect. 

However, there is ongoing debate about whether \acp{llm} truly engage in human-like internal mechanisms with genuine understanding~\cite{wei2022chain,webb2023emergent,mitchell2023debate,gurneelanguage}, or whether they are merely leveraging statistical patterns without attaining the functional cognitive competence of human brains~\cite{mitchell2023debate,mahowald2024dissociating}. On one hand, \acp{llm} can accomplish some logical reasoning tasks; on the other hand, they still struggle on even simple reasoning problems that lie outside of their training distribution, suggesting an absence of robust generalization~\cite{saparov2023testing,chen2024premise,malek2025frontier}. This controversy ties into the well-known dissociation between language and thought in the human brain~\cite{fedorenko2024language}. It also raises the question of whether learning from next-token prediction alone is sufficient to yield complex and generalizable reasoning or if additional architectural modules or training signals are needed to reach human-like competence~\cite{mahowald2024dissociating}. Addressing these questions requires examining not just the models' outputs, but also their internal mechanisms such as neural representations.

For insight into these issues, recent NeuroAI studies have compared the internal representations of \ac{ai} models with human brain-activity patterns using neural predictivity metrics (often termed ``brain scores'')~\cite{schrimpf2018brain,yamins2014performance,schrimpf2021neural}. For example, computer vision models produce representations that closely mirror those in visual cortices~\cite{yamins2014performance,zhuang2021unsupervised,wang2023better}; language and speech models display activity patterns paralleling the brain's core language network~\cite{schrimpf2021neural,caucheteux2022brains,goldstein2022shared,kumar2024shared,goldstein2025unified}; and multimodal models align with object representations in high-level visual regions~\cite{du2025human}. Moreover, there is evidence that \acp{llm} can drive and suppress activity in human language network~\cite{tuckute2024driving}, and incorporating neural signals into model training can improve the robustness of computer vision models~\cite{li2019learning,dapelloaligning} and enhance the semantic understanding of speech models~\cite{moussa2025improving}.

Despite these advances, it remains unclear how \ac{llm} representations relate to the brain regions that support complex reasoning, especially given that distinct neural substrates underlie language and logical reasoning in humans~\cite{fedorenko2024language,prado2011brain}. Furthermore, a critical open question is whether human neural activation patterns could be used to directly enhance \ac{ai} models' performance on such cognitive tasks, specifically, to augment the kind of deductive reasoning ability with which \acp{llm} still struggle---failing even on out-of-distribution yet simple problems~\cite{saparov2023testing,chen2024premise,malek2025frontier}. Examining these issues at the representation level beyond the models' language outputs could offer a new perspective for both understanding and improving \acp{llm}.

In this study, focusing on deductive reasoning (\cref{fig:overview}), we provide evidence that \acp{llm} partially align with human brain activity in reasoning-related regions, and that this alignment can be harnessed to improve the models' reasoning performance using neural signals. First, we show that the internal representations of open-source \acp{llm} are partially aligned with human neural activations in brain regions associated with deductive reasoning. Using a task-specific \ac{fmri} dataset, we find that these \ac{llm} representations can explain roughly 76\% of the explainable variance in aggregate neural responses across deductive reasoning regions~\cite{prado2011brain}, whereas their predictivity within each specific reasoning type is lower ($\sim$27\%). Notably, this model-brain correspondence is far above the control level (untrained models' representations) and is selective for the brain's reasoning network over other regions such as the \ac{md} and core language networks. These results suggest that \acp{llm} are partially emulating cognitive mechanisms beyond those of language processing, although their reasoning capabilities and the alignment remain imperfect.

Building on this partial alignment, we introduce an approach to actually enhance \ac{llm} reasoning using human neural data. In essence, we guide the model's internal representations---that will produce incorrect responses---toward directions informed by the joint structure of model and brain representations, viewing neural data as a tool for performance improvement rather than treating global model--brain alignment as the ultimate goal. We implement this idea in two ways: through \ac{nari} applied at inference time, and \ac{narf} that adjusts the model's parameters. Our experiments show that neural guidance can effectively redirect an \ac{llm}'s answers from incorrect to correct on the fly, and that these interventions generalize (in part) to novel questions and to other forms of reasoning. Moreover, by fine-tuning the model with representational guidance (\ac{narf}), we can inject the representational neural knowledge into its parameters, enabling the model to generalize to different types of deductive reasoning beyond the training data. Notably, this approach is distinct from conventional language-behaviour supervision; we find that neural guidance provides complementary performance gains and yields more refined internal reasoning trajectories. We validate our approach on \acp{llm} ranging from 1.5 billion to 70 billion parameters, achieving performance improvements even for models that initially exceeded human accuracy. To our knowledge, this work represents a pioneering demonstration that cognitive brain signals can directly improve an \ac{ai} model's task performance.

Taken together, our findings indicate that \acp{llm} and the human brain share partially overlapping mechanisms for deductive reasoning, evident both in correlational representational alignment and in the functional gains achieved when neural signals are used to guide models. This work introduces a brain-signal-guided pathway to advance \acp{llm} by integrating the structure of human cognitive neural data, complementary to behavioural supervision, marking a significant step towards cognitively principled and capable \ac{ai} systems.

\section{Results}

\subsection{\texorpdfstring{\acp{llm}}{} partially align with human neural activity during deductive reasoning}\label{sec:result:llm-brain-connection}

\acp{llm} are trained exclusively on textual data, yet human logical reasoning recruits neural circuits that are largely distinct from language-processing regions. This dichotomy motivates a key question: can \acp{llm} exhibit human-like brain representation patterns beyond those associated with language? To probe this, we focus on deductive reasoning, which engages a fronto-parietal reasoning network largely separable from classical language areas~\cite{prado2011brain}.

We leverage an open-source \ac{fmri} dataset (OpenNeuro Dataset ds003076) in which ten healthy adults (ages 19-30) solved 70 deductive logic problems spanning two types, syllogistic and transitive reasoning, with a comprehensive factorial design (Methods). The dataset leverages pseudowords that minimize semantic and lexical confounds, providing a cleaner basis for comparing neural representations of reasoning with language models. On each trial, three premises are presented sequentially, followed by a conclusion requiring a binary validity judgement (``True''/``False''). In parallel, we evaluate ten open-source \acp{llm} spanning 1.5B-72B parameters from different families: Qwen2-1.5B-Instruct, Qwen2-7B-Instruct, Qwen2-72B-Instruct~\cite{yang2024qwen2}, Qwen3-4B~\cite{yang2025qwen3}, Llama2-7B-Chat~\cite{touvron2023llama}, Llama3-8B-Instruct, Llama3.3-70B-Instruct~\cite{dubey2024llama}, Mistral-7B-Instruct~\cite{jiang2023mistral}, Phi-4-mini-Instruct~\cite{abouelenin2025phi}, and Gemma2-9B-it~\cite{team2024gemma} (instruction-tuned models; model names shortened in figures; base models show similar trends, \cref{supp:sec:base-model}). Each model receives the same stimuli as in the \ac{fmri} protocol (three premises, then the conclusion; Methods, \cref{fig:overview}a) and judges the conclusion's validity. Model accuracies range from 71.4\% (Llama2-7B-Chat) to 97.1\% (Llama3.3-70B-Instruct), bracketing the human average (86.4\%), thereby establishing a meaningful basis for model-brain comparison.

We quantify model-brain correspondence using neural predictivity, defined as the fraction of explainable variance in extracted \ac{fmri} responses accounted for by functionally localized model units~\cite{alkhamissi2025llm,alkhamissi2025language} (predictivity-ceiling normalized; Methods; \cref{fig:brain-score}a). For each model, we first localize model units responsive to deductive reasoning via contrasting~\cite{alkhamissi2025llm,alkhamissi2025language} from each transformer layer's output and the attention output~\cite{kumar2024shared} at the last token, and then extract representations from these units for each problem. We also calculate neural predictivity for each layer of the model. We then fit cross-validated ridge regression models to predict voxel-wise \ac{fmri} responses from these representations and calculate the Pearson correlation (Methods). Predictivity is assessed in reasoning ROIs~\cite{prado2011brain} and compared against \ac{md} and core language networks (Methods).

Across all trials (syllogistic + transitive), \acp{llm} capture $\sim$76\% of explainable variance in reasoning regions (\cref{fig:brain-score}b). Predictivity is significantly higher in reasoning and \ac{md} regions than in language regions (two-sided paired t-tests across models; $p<0.001$ for reasoning $>$ language; $p<0.001$ for \ac{md} $>$ language; \cref{fig:brain-score}b). Anatomically, these brain regions exhibit distinct functional profiles (\cref{fig:brain-score}c): deductive reasoning regions encompass voxels responsive to general, syllogistic, and transitive reasoning~\cite{prado2011brain}; \ac{md} regions support demanding cognitive tasks, including logical reasoning~\cite{duncan2010multiple}; and language networks specialize in linguistic processing.

Analyzing reasoning types separately, ceiling-normalized predictivity drops to $\sim$27\% for each type (\cref{fig:brain-score}d), yet remains: (i) significantly higher than untrained models (random model weights with trained tokenizer; two-sided paired t-test across models, $p<0.001$ for all comparisons; \cref{fig:brain-score}e); and (ii) regionally specific (deductive reasoning $\geq$ language $\geq$ \ac{md}; two-sided paired t-test across models, for syllogistic reasoning: $p = 0.001$ for deductive reasoning vs. \ac{md} and $p = 0.009$ for deductive reasoning vs. language; for transitive reasoning: $p = 0.01$ for deductive reasoning vs. \ac{md}; \cref{fig:brain-score}f). This pattern is consistent with \ac{md} reflecting domain-general demands across problems, while the reasoning network carries more type-specific structures.

Visualization through \ac{pca} reveals that both \ac{llm} and \ac{fmri} representations partially separate syllogistic from transitive problems (\cref{fig:brain-score}g), reconciling high cross-type predictivity with lower within-type predictivity. Middle layers and attention outputs of models generally show higher alignment (\cref{fig:brain-score}h, \cref{sup:fig:different-layer}).

Together, our results reveal that \acp{llm}' internal representations partially align with human brain activity in deductive reasoning regions, albeit with notable differences. Among all brain areas examined, the model representations show closer correspondence to the deductive reasoning network than to \ac{md} or language regions. This regional specificity suggests that \acp{llm} do capture certain reasoning-related neural patterns beyond pure linguistic processing. 

\subsection{Improving \texorpdfstring{\acp{llm}}{} with human neural activations}\label{sec:result:neuroscience4ai}

Given the partial alignment between model and brain representations, a natural question arises: can we leverage human brain signals to further guide and improve a model's reasoning representations? This is particularly intriguing in light of our previous findings: because \acp{llm} already partially exhibit human-like patterns of brain activity during deductive reasoning, introducing appropriate human brain signals into the model might selectively amplify its deductive reasoning capabilities.

To investigate this possibility, we explore a brain-guided approach that steers \ac{llm} internal representations using human neural activations, moving beyond traditional behaviour-level supervision. We introduce two complementary methods---\ac{nari} and \ac{narf}---that directly incorporate task-\ac{fmri} signals from reasoning tasks into the model's latent representations. 

\begin{figure*}[t!]
    \centering
    \includegraphics[width=\linewidth]{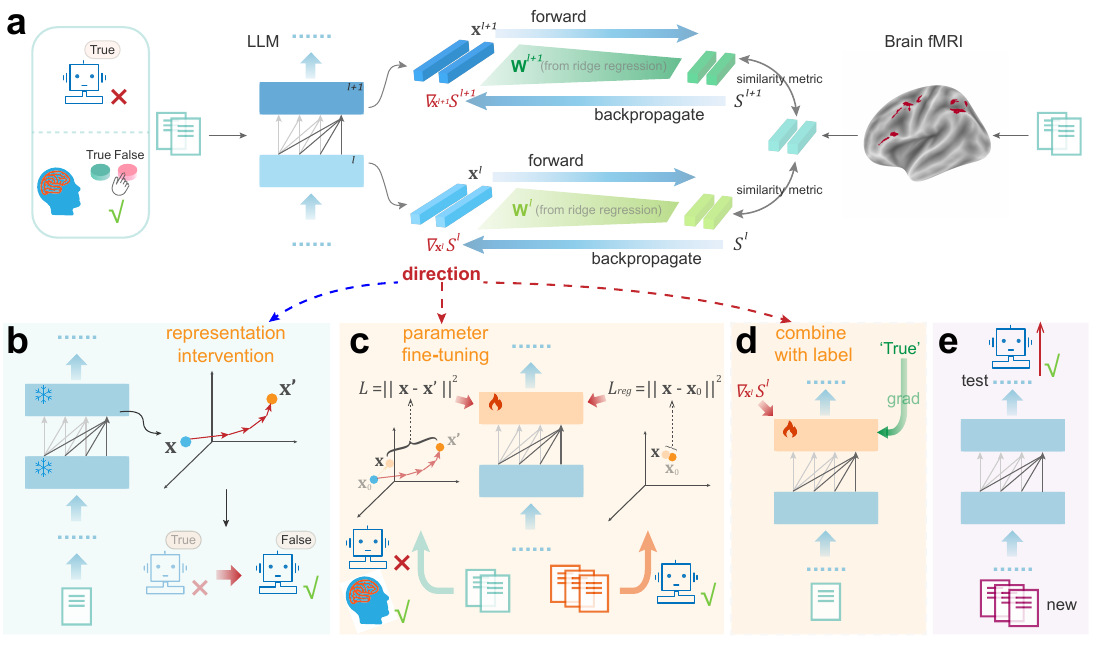}
    \caption{\textbf{Illustration of the Methods to Steer \acp{llm} with Human Neural Activations.} (a) We induce directions at the model representation space of each layer for problems with incorrect responses. We calculate gradients of model representations $\mathbf{x}$ to maximize a similarity metric between \ac{fmri} signals and  $\mathbf{x}$ transformed to the \ac{fmri} space via weights from ridge regression, which encodes the structures of model and \ac{fmri} representations. (b) We apply these directions to modify representations through inference-time intervention without parameter updates, correcting model outputs. (c) The directions that induce successful intervention are used to fine-tune model parameters, injecting representational knowledge into the parameters. (d) The neural-derived directions complement output language supervision during fine-tuning through the combination of objectives. (e) The models improve test performance on new deductive reasoning problems.}
    \label{fig:illustration}
\end{figure*}

At the core of our approach is a representation-level, brain-signal-guided objective that yields a structure-derived direction in the model's latent space (\cref{fig:illustration}). For each layer $l$, we first learn a linear encoder that maps an \ac{llm}'s hidden states to \ac{fmri} space by fitting a ridge regression $\hat{y}=W_l x_l + b_l$ to predict the \ac{fmri} response $y$ for deductive reasoning questions given \ac{llm}'s representation $x_l\in\mathbb{R}^d$ (Methods; \cref{sup:sec:explain-narinarf}). We then define a similarity $S(W_l x_l, y)$ and compute its gradient with respect to the representation while excluding the intercept $b_l$:

\vspace{-4mm}
$$ g_l(x_l) = \nabla_{x_l} S(W_l x_l, y) = {W_l}^\top \nabla_{W_l x_l}S,$$
which is explicitly induced by the joint structure of model and brain representations, and includes a structure-mediated systematic shift between these representations (Methods; \cref{sup:sec:explain-narinarf}). The gradients induce directions and are estimated per reasoning type (syllogistic or transitive), providing a representational steering direction without language behaviour labels.

We use the gradient-derived direction in two complementary ways. (i) Inference-time intervention (\ac{nari}): we nudge the representation by the direction within a bounded range to correct erroneous trajectories during inference (\cref{fig:illustration}b). (ii) Parameter fine-tuning (\ac{narf}): we internalize the representational neural knowledge by optimizing parameters so that intermediate representations either maximize $S(W_l x_l(\theta), y)$ or approach a \ac{nari}-shifted target (\cref{fig:illustration}c). Finally, we further show that this neural, representation-level objective is complementary to output language supervision by studying a hybrid objective combining both of them (\cref{fig:illustration}d). Implementation details are in Methods, \cref{sup:sec:explain-narinarf}, and \cref{supp:sec:implementation-details}.

\subsubsection{Human neural activations guide effective representation intervention directions}

We first validate our approach through representation intervention experiments. We apply \ac{nari} to the attention-module outputs in middle transformer layers (1/4-3/4 depth) at inference (Methods). 
This layer selection is motivated by (i) the role of attention in modeling token relationships~\cite{vaswani2017attention} critical for deductive reasoning, (ii) the tendency of intermediate layers to encode abstract features~\cite{templeton2024scaling,hojerimproving}, and (iii) their higher neural predictivity in our analysis (\cref{fig:brain-score}h, \cref{sup:fig:different-layer}).

Our intervention paradigm (\cref{fig:intervention}a-c) comprises three components: (1) Targeted intervention: evaluate whether a candidate direction can effectively correct the model's answer within a bounded perturbation range (\cref{fig:intervention}a, Methods; extended explanation in \cref{sup:sec:extended-explanation}). Smaller admissible ranges indicate more precise directions. (2) Subject-wise aggregation: aggregate success across all 10 human subjects (\cref{fig:intervention}b, Methods). (3) Generalized direction: average successful subject-specific intervention direction to obtain a general direction applicable to new problems (\cref{fig:intervention}c, Methods). We compare \ac{nari} to two baselines: (1) Random Signals: replace true \ac{fmri} signals with random data, preserving only the model's representational structure, and (2) Random Directions: sampled intervention vectors, with the number matching the subject count (Methods).

We consider six models (Qwen2-1.5B/7B-Instruct, Mistral-7B-Instruct, Llama2-7B-Chat, Llama3-8B-Instruct, Phi-4-mini-Instruct) that have errors in both syllogistic and transitive questions. 
The results of six models demonstrate that \ac{nari} achieves a 100\% success rate in flipping initially incorrect answers to correct ones, significantly outperforming both baselines across perturbation ranges (\cref{fig:intervention}d; individual results in \cref{sup:fig:intervention-each-model}). Model representational structure alone (Random Signals baseline) shows limited efficacy, underscoring the importance of human neural structure beyond the model's intrinsic geometry.

We observe variability across the representational structures of human subjects and models (\cref{sup:fig:model-fmri-structure}) and the success rate among subjects (\cref{supp:sec:analysis-nari}, \cref{sup:fig:intervention-analysis}c-d), indicating that intervention effectiveness depends on the model-subject coupling; while generally higher-performing subjects yield higher success rate (\cref{sup:fig:intervention-analysis}d). Meanwhile, region specificity holds: signals from deductive reasoning regions yield higher success than \ac{md} or language regions (\cref{fig:intervention}e). Excluding the intercept term when computing gradients, which introduces a systematic shift in the direction, improves direction quality (\cref{fig:intervention}f; Methods, \cref{sup:sec:explain-narinarf}; \cref{supp:sec:analysis-nari}, \cref{sup:fig:intervention-analysis}a-b).

\ac{pca} visualizations illustrate the mechanism (\cref{fig:intervention}g): before intervention, latent representations for ``True'' and ``False'' answers are often mixed; after \ac{nari}, representations of previously incorrect cases shift toward the correct-answer subspace, through the directions that emerged from the interaction between model and \ac{fmri} representational structures. 

Further, the general direction transfers to new, out-of-set problems (\cref{fig:intervention}h; Methods), improving multiple models (Qwen2-1.5B-Instruct, Llama2-7B-Chat, Llama3-8B-Instruct, Phi-4-mini-Instruct), while random directions fail. Two models (Qwen2-7B-Instruct, Mistral-7B-Instruct) show no improvement, likely due to too few incorrect examples to estimate reliable directions and the potential instability of representation steering~\cite{tan2024analysing,da2025steering}. 

Moreover, given the recent advancement in language-based ``thinking'' models~\cite{guo2025deepseek} with chain-of-thought, we also explore the extension of \ac{nari} to a thinking model (DeepSeek-R1-Distill-Qwen-1.5B). We find that \ac{nari} can successfully modify the language-based reasoning process toward correct conclusions (\cref{fig:intervention}i), where the structure of human neural activations is important (\cref{fig:intervention}j), and the general direction can transfer improvements to new problems (\cref{fig:intervention}k).

These results demonstrate that human neural signals can induce effective representational guidance for LLMs, and this neural guidance can transfer, suggesting a general knowledge contained in the neural-induced representational directions.

\subsubsection{Fine-tuning models generalizes improvements to new deductive reasoning problems}

\begin{figure*}[t!]
    \centering
    \includegraphics[width=\linewidth]{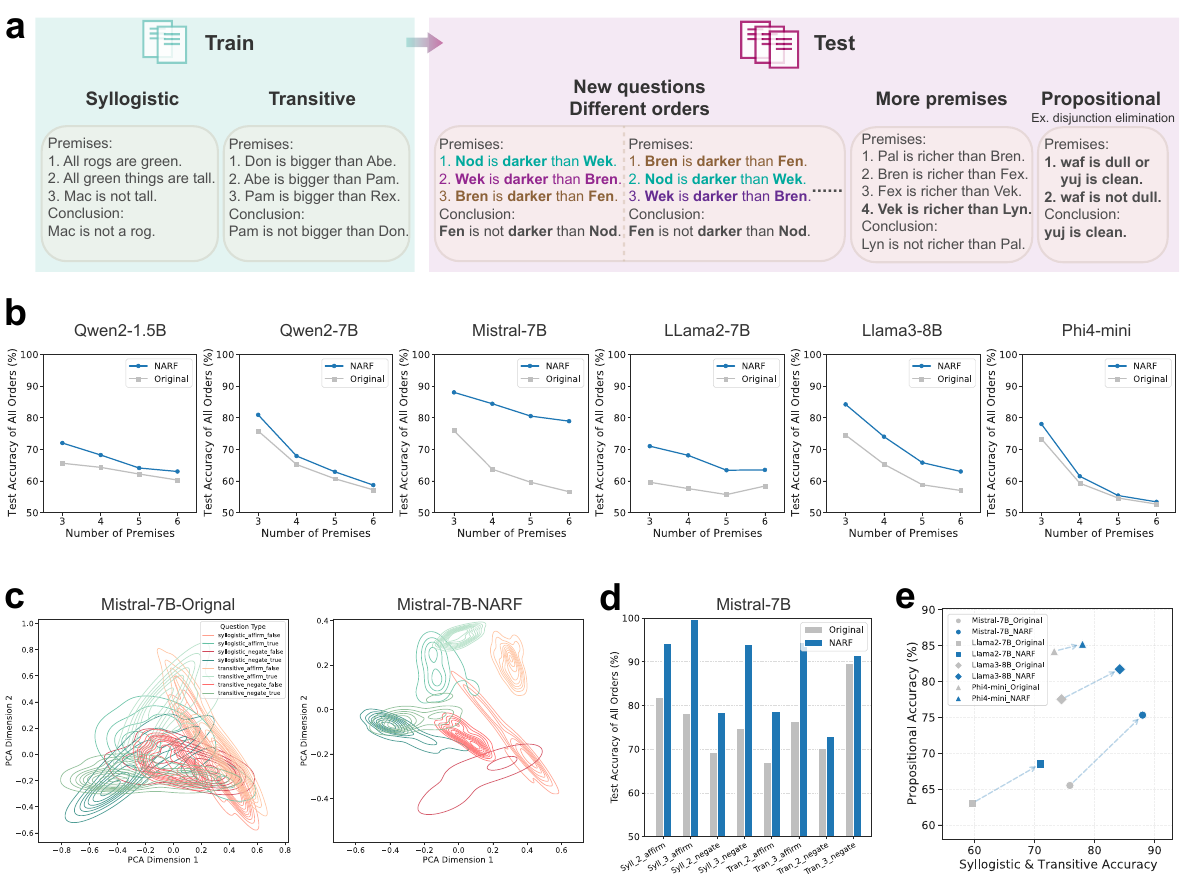}
    \caption{\textbf{Results of Parameter Fine-tuning Experiments.} (a) Illustration of the training and testing settings. Training data includes syllogistic and transitive reasoning problems from the \ac{fmri} dataset, while we test models on new stimuli with varying premise orders, extended premise numbers, and different reasoning types. (b) Persistent testing accuracy improvement of various models on new data across different premise configurations, demonstrating robust generalization. (c) Kernel-density-estimation-based visualization of the model representations (attention output of the 3/4-th layer) on testing data reveals increased separation between problems with ``True'' and ``False'' labels in the representation space after fine-tuning. (d) Testing results on different subtypes of reasoning problems demonstrate consistent improvements (full results in \cref{sup:fig:finetune-subtype}). (e) Test results on propositional reasoning problems. Improvements of the model's behaviour generalize to other types of reasoning.}
    \label{fig:finetune1}
\end{figure*}

We next test whether this representation-level knowledge can be internalized via parameter updates. \ac{narf} fine-tunes model parameters so that intermediate representations approach brain-guided targets (\eg, successful \ac{nari} representations; Methods). Models are trained on the syllogistic/transitive problems from the \ac{fmri} dataset and evaluated on newly generated tests (\cref{fig:finetune1}a), including questions with permuted premise orders, increased numbers of premises, and propositional reasoning (a different deductive reasoning type with distinct neural basis from syllogistic and transitive problems~\cite{prado2011brain}). The generated test data also utilize pseudowords to minimize semantic confounds, and evaluate the generalization under distribution shifts.

The \ac{narf} fine-tuned models yield robust gains across all evaluation scenarios. Accuracy improves for permuted-order and more-premise problems (\cref{fig:finetune1}b), indicating true generalization rather than mere pattern matching. After fine-tuning, latent representations better separate correct from incorrect logic on test problems (\cref{fig:finetune1}c), consistent with accuracy gains. Improvements persist across fine-grained reasoning subtypes (\cref{fig:finetune1}d; full results in \cref{supp:sec:narf-subtype} and \cref{sup:fig:finetune-subtype}). Further, the benefits extend to propositional reasoning for most models (\cref{fig:finetune1}e), except for two Qwen models whose original accuracy is either below random guess or near-ceiling. 

Overall, representational enhancements learned via \ac{narf} transfer to distinct questions and across deductive types, indicating that neural guidance can be internalized into parameters and offering a representation-level route distinct from behaviour-level supervision for parameter training.

\subsubsection{Neural activation guided representation fine-tuning complements language supervision}

Since \ac{narf} targets intermediate representations rather than final outputs, we further study its synergy with language supervision and show the complementary benefits. While standard language supervision optimizes final output tokens (\eg, cross-entropy on ``True''/``False''), \ac{narf} simultaneously encourages intermediate representations toward a higher similarity metric with human neural activations (\cref{fig:illustration}d, Methods).

\begin{figure*}[t!]
    \centering
    \includegraphics[width=\linewidth]{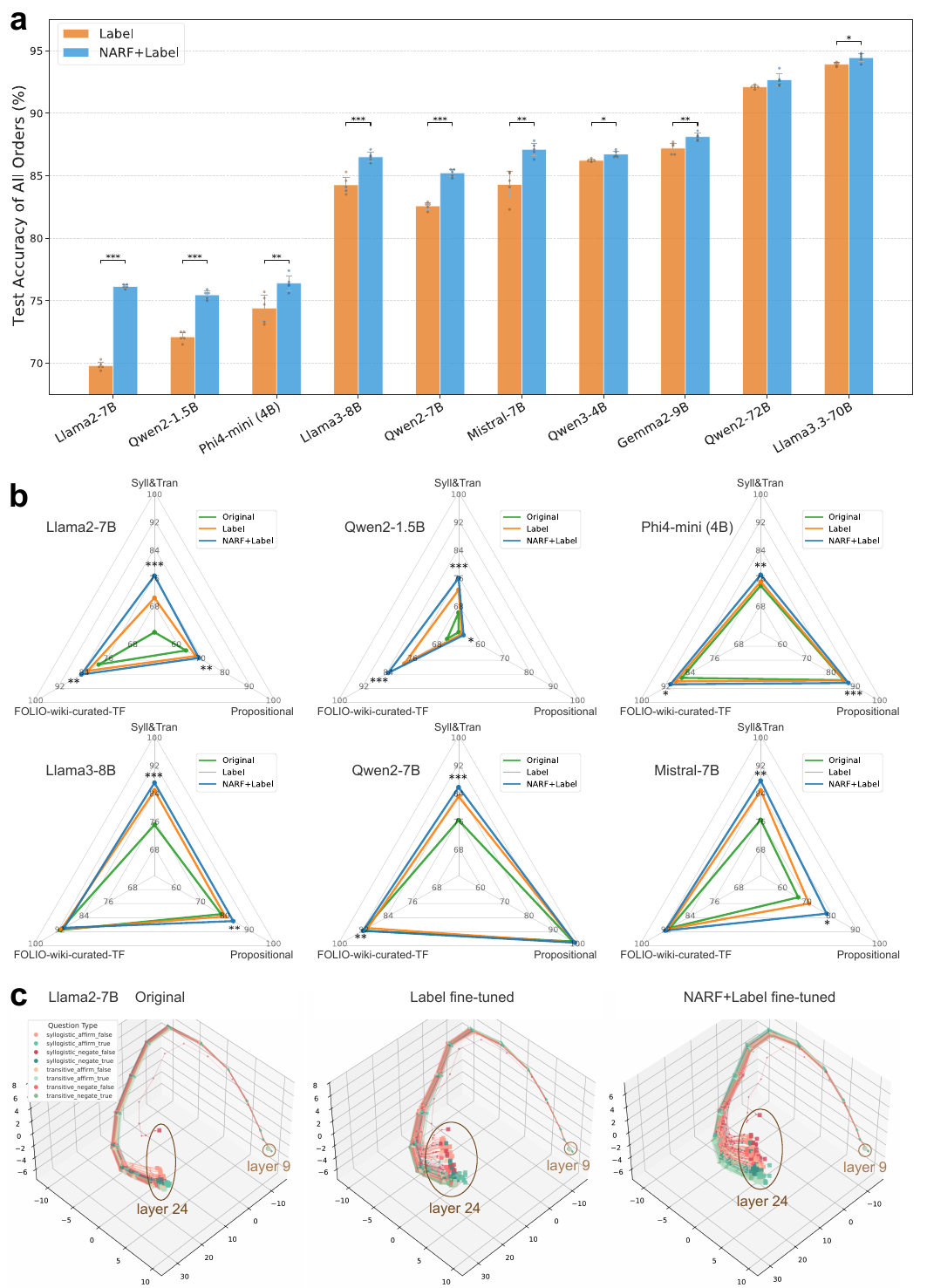}
    \caption{\textbf{Results of Combining \ac{narf} with Language Supervision.} Statistical tests are reported based on two-sided paired t-test over five runs of experiments under same random seeds. (a) Testing accuracies (mean$\pm$std) of various models (1.5B--70B parameters) for language-label-only and hybrid (\ac{narf}+language labels) fine-tuning, ordered based on models' original test accuracy. (b) Testing performance (mean) comparison on diverse reasoning data. Improvements on syllogistic and transitive problems generalize to propositional reasoning and the FOLIO dataset with first-order logic in natural language. (c) Visualization of the layer-wise reasoning trajectories (layer outputs from the 1/4-th layer to the 3/4-th layer) in the representation space shows enhanced separation between paths for correct and incorrect logics after hybrid fine-tuning compared to the original and label-only fine-tuned models.}
    \label{fig:finetune2}
\end{figure*}

We evaluate the approach across 10 \acp{llm} spanning diverse scales and from different families. The results show that augmenting language supervision with \ac{narf} yields consistent, statistically significant gains (\cref{fig:finetune2}a), with 2.2\% average accuracy gain in average (range 0.5–6.4\%) on test problems with all premise-order permutations. Ablation analyses confirm that the gains arise from the joint influence of model and human neural representational structures, and human fMRI targets outperform label-based representations (\cref{supp:sec:ablation-narflabel}; \cref{sup:fig:narflabel-ablation}).

The hybrid approach further enhances generalization to different reasoning types, including generated propositional problems and the FOLIO dataset~\cite{han2024folio} (\cref{fig:finetune2}b). For instance, on propositional problems, Mistral-7B-Instruct is improved by 13.2\% considering the max performance under five runs (83.7\%) over the max performance of language-only fine-tuning (70.5\%). Representational analyses suggest a potential mechanism: language supervision mainly shapes final-layer outputs, whereas NARF induces more discriminative reasoning trajectories throughout the network (\cref{fig:finetune2}c; additional results in \cref{sup:fig:more-reasoning-path}), yielding clearer separation of correct vs. incorrect logics and supporting superior generalization performance.

In summary, the results demonstrate that human neural activations provide a distinct representation-level learning signal that complements conventional behaviour-level supervision, encouraging generalization performance. Neural-guided fine-tuning enables models to refine implicit, intermediate trajectories while remaining fully compatible with standard language-based training paradigms.

\paragraph{Extension to other neural datasets}
We further validate our method on another independent dataset, HCP Relational Processing Task from the Human Connectome Project~\cite{van2013wu} (\cref{sup:sec:hcp}, \cref{sup:fig:hcp}). This task probes relational reasoning by comparing relations from two pairs of images, and we adapt it to large language models by presenting language descriptions of the attributes of the images to the models (\cref{sup:fig:hcp}a-b). This dataset differs from the deductive reasoning dataset in both task and stimulus modality, thereby providing a meaningful external validation of the overall pipeline.

The results show that human neural activations from this dataset can also induce effective representation intervention directions, significantly outperforming both baselines (model representational structure alone / random directions) across perturbation ranges (\cref{sup:fig:hcp}c), and the general direction of NARI can transfer to new, out-of-set problems (\cref{sup:fig:hcp}d). At the same time, NARF complements language supervision with consistent gains (\cref{sup:fig:hcp}f). Full results and details can be found in \cref{sup:sec:hcp}.

The results show that our pipeline is not limited to one specific experimental design. Meanwhile, they show that our method can extend to other related reasoning tasks with different stimulus modalities, suggesting a promising way to incorporate more potential neural signal data.

\section{Discussion}

The rapid advancement of \ac{ai} has raised fundamental questions about the relationship between artificial and human intelligence: To what extent do \ac{ai} models (\eg, \acp{llm}) replicate human cognitive processes, and how can neuroscience inform their improvement? Focusing on deductive reasoning, a critical component of logical thinking, we demonstrate that \acp{llm} exhibit partial alignment with human brain activity, while divergences remain. Crucially, we establish that task-based \ac{fmri} signals can directly guide improvements of \ac{ai} models in higher-level cognitive tasks at the representational level, offering a different paradigm for model enhancement. 

Neuroscientific evidence suggests a potential dissociation between language and thought in humans, with distinct neural networks supporting linguistic processing and other cognitive demands~\cite{fedorenko2024language}. Therefore, it remains debatable whether language models primarily capture formal linguistic competence~\cite{mahowald2024dissociating,mitchell2023debate} or if they can capture functional competence and develop an internal world model~\cite{liemergent,gurneelanguage} with language as a carrier of thought. Our findings demonstrate that these models can partially approximate human-like cognitive processes such as deductive reasoning, aligning with recent findings to capture human cognition with \ac{ai} models~\cite{binz2025foundation}. The observed alignment with deductive reasoning networks, as well as the partially shared representation space that enables direct guidance, supports the hypothesis that \acp{llm} may develop internal representations that go beyond surface-level linguistic patterns, suggesting that language models can extend beyond simulating human language networks and potentially resemble other cognitive systems when trained on diverse data. At the same time, there still exists divergence between artificial and biological systems, and we further demonstrate that human neural activations can guide the improvement of \acp{llm}.

Reasoning has emerged as a critical frontier in \ac{ai} and \ac{llm} research, with massive recent advances such as DeepSeek-R1~\cite{guo2025deepseek} and thinking models. Despite the achievements, frontier \acp{llm} are still shown to struggle on simple reasoning tasks, highlighting the out-of-distribution generalization problem~\cite{malek2025frontier}. Our work provides an alternative perspective on learning intermediate representations directly from human neural signals rather than solely through language behaviour, offering a different possibility for ``process supervision''. This approach doesn't compete with, but rather complements language behaviour supervision (\cref{fig:finetune2}; \cref{sup:sec:combine-synthetic}). We mainly focus on the basic deductive reasoning problems and implicit reasoning in \acp{llm}, in accordance with the available fMRI data (most response times around 1-3s~\cite{lytle2020children}, indicating fast thinking is often utilized), while our demonstration on chain-of-thought models reveals promising directions for future work on more complex cognitive processes with slow thinking. 
As fMRI signals are slow hemodynamic responses that may not reflect fast dynamical processes of chain-of-thought reasoning, we only consider aggregated representations, while future investigation on modalities with higher temporal resolution such as MEG or EEG may overcome this limitation by tracking fast intermediate steps for potentially more fine-grained guidance.

Given the basic reasoning problems in the fMRI dataset, we mainly target mid-scale \acp{llm} (1.5B-70B parameters) that still exhibit measurable errors, allowing us to rigorously verify the feasibility of neural-guided improvement. On the other hand, it also presents the possibility to improve efficient \acp{llm} beyond increasing parameters and language data, given the better generalization performance of neural-guided enhancement compared with language-only training. 
Future work can further investigate more complex cognitive tasks and even larger models as richer neural datasets become available.

Our methodology represents a significant departure from previous neuro-inspired \ac{ai} approaches that focused on mimicking low-level neuronal dynamics~\cite{roy2019towards,rao2022long,hasani2022closed} or implementing brain-derived rules or architectures~\cite{kudithipudi2022biological,wang2023incorporating,xiaohebbian,achterberg2023spatially}. Instead, we demonstrate that the internal representations in \ac{ai} models for high-level cognitive tasks can be directly improved by human neural data. This aligns with emerging evidence of convergent representations across biological and artificial systems~\cite{huhposition}, suggesting new opportunities for neuroscience to inform \ac{ai} development. Some previous work explored brain-signal-induced improvements for adversarial robustness in computer vision models~\cite{li2019learning,dapelloaligning}, semantic understanding in speech models~\cite{moussa2025improving}, or encouraging the similarity between models and brains~\cite{schwartz2019inducing}, while our study advances the field to show how brain signals for cognitive tasks can directly and functionally enhance \ac{llm} capabilities. This initiates the road to improve cognitive abilities of \ac{ai} models by human neural representations apart from behaviour supervision~\cite{binz2025foundation}, and provides opportunities for other cognitive capabilities, such as theory of mind that has a specialized brain network~\cite{fedorenko2024language} and potential representational alignment considering performance~\cite{strachan2024testing} and representation~\cite{zhulanguage} of \acp{llm}.

In conclusion, our work establishes both the partial neurocognitive alignment between \acp{llm} and human brains and the viability of using human brain signals (\eg, task \ac{fmri}) to improve \ac{ai} models for cognitive tasks such as deductive reasoning. These findings open numerous research avenues, from investigating more complex cognitive tasks to developing hybrid training paradigms that combine neural guidance with conventional approaches. We anticipate this work will not only catalyze deeper understanding of connections between artificial and biological intelligence, but also provide concrete methods for building more human-like \ac{ai} systems through neuroscience-inspired approaches.

\section{Methods}\label{sec:methods}

\subsection{Datasets}

\paragraph{fMRI dataset} 
We leverage an open-source \ac{fmri} dataset: OpenNeuro Dataset ds003076 (Lytle MN, Prado J \& Booth JR (2020). Brain correlates of deductive reasoning in adults. OpenNeuro. \url{https://openneuro.org/datasets/ds003076}), with a corresponding data descriptor~\cite{lytle2020children}. 
The dataset contains recordings from 17 participants (age range: 19--30 years). After excluding 6 participants with missing behavioural data in one of the four sessions or accuracy below chance level in any session, and 1 participant whose MRI image could not be segmented and normalized well in our preprocessing, our final analysis comprises 10 subjects (mean age: $25.3\pm 3.4$ years). 

\textit{Task description}\quad In experiments of this \ac{fmri} dataset, participants performed two types of deductive reasoning tasks during \ac{fmri} acquisition: 1) syllogistic reasoning: set-inclusion problems describing a series of relationships (adjective) among three classes (monosyllabic pseudoword), with three premises and a conclusion, \eg, ``Premises: \textit{All blons are pink; All pink things are young; Ken is a blon;} Conclusion: \textit{Ken is pink}''; and 2) transitive reasoning: linear ordering problems with comparative adjectives between imaginary characters (syllable name), \eg, ``Premises: \textit{Ned is slower than Max; Max is slower than Sam; Sam is slower than Rob;} Conclusion: \textit{Sam is slower than Ned}''. In each experimental trial, stimuli (three premises and one conclusion) were presented sequentially (2s interval per item) with the next sentence appearing below the last, and participants were given 6s response windows to press one of two buttons to judge the validity of the conclusion (\cref{fig:overview}b). The dataset originally contained 72 problems (36 per task) across four conditions: 1) true affirmative (18), 2) false affirmative (6), 3) true negation (6), and 4) false negation (6), where true/false refers to the correctness of the conclusion while affirmative/negation describes the conclusion (\eg, \textit{Ken is pink} or \textit{Ken is not pink}), and each condition further contains two types requiring 2 or 3 premises to make the judgement. After a manual check, we remove 2 contradictory transitive problems (\eg, A is taller than B, B is taller than C, C is taller than A), and analyze 70 problems in total.

\textit{Data preprocessing}\quad We use MATLAB, SPM12, and GLMSingle~\cite{prince2022improving} to preprocess \ac{fmri} data and extract neural responses to each reasoning problem (details in \cref{supp:sec:fmri-preprocess}). We further extract voxels from the estimated response based on locations of deductive reasoning regions identified in the meta-analysis~\cite{prado2011brain}, involving left \ac{ifg}, left \ac{mefg}, bilateral \ac{mfg}, bilateral \ac{pg}, bilateral \ac{ppc}, and left \ac{bg}. We also consider language and \ac{md} regions (available from \url{https://www.evlab.mit.edu/resources}, which are widely used in previous works~\cite{schrimpf2021neural}). We imitate participant-specific functional localization~\cite{fedorenko2010new} to specify locations for each human subject. The extracted voxels are considered vectors of brain representations (more details in \cref{supp:sec:fmri-preprocess}).

\paragraph{Generated datasets}
In addition to deductive reasoning problems in the \ac{fmri} dataset, we generate additional test datasets to evaluate model deductive reasoning abilities. We follow the \ac{fmri} dataset to use random pseudowords in order to reduce the possibility that the training data of \acp{llm} contain them.

(1) \textit{Syllogistic and transitive problems}. We create new problems matching the structure of problems in the \ac{fmri} dataset, consisting of several premises and a conclusion, but with 1) different monosyllabic pseudowords, syllable names, and adjectives, which are randomly generated or chosen from a list; 2) balanced conditions with the same number of problems for all conditions (unlike the original fMRI distribution); and 3) two variants: the similar 3-premise setting with 8 conditions (true/false $\times$ affirmative/negation $\times$ 2/3 required premises), and the more-premise setting (4--6 premises) with 4 conditions (true/false $\times$ affirmative/negation). 

The validation dataset only considers the 3-premise setting, with 10 problems per condition. The test dataset with 3/4/5/6 premises consists of 100/50/20/10 problems for each condition, respectively. We enumerate all possible permutation orders of premises during validation and testing to reduce the influence of premise order~\cite{chen2024premise}; therefore, the total number of validation/testing problems will be 6/24/120/720 times expanded (more details in \cref{supp:sec:generated-data}).

(2) \textit{Propositional problems}. Propositional reasoning is a different reasoning type from syllogistic and transitive ones, also with a distinct neural basis in the brain~\cite{prado2011brain}. We generate propositional problems with three classical types: 1) modus ponens (\textit{If P then Q; P; therefore Q}), 2) modus tollens (\textit{If P then Q; not Q; therefore not P}), and 3) disjunction elimination (\textit{P or Q; not P; therefore Q})~\cite{prado2011brain}, each with true or false conclusions. Each condition contains 100 problems, resulting in 600 problems in total (more details in \cref{supp:sec:generated-data}).

\paragraph{FOLIO} FOLIO dataset~\cite{han2024folio} is a first-order logical inference dataset for reasoning in natural language. Each question consists of multiple premises and a hypothesis. We use the FOLIO-wiki-curated version~\cite{zhang2023cumulative} that removes problematic instances, and we only consider questions with ``True'' or ``False'' answers to align with our setting, with a total number of 315 questions (more details in \cref{supp:sec:folio}).

\subsection{Models}

\begin{figure}[t!]
    \centering
    \includegraphics[width=\linewidth]{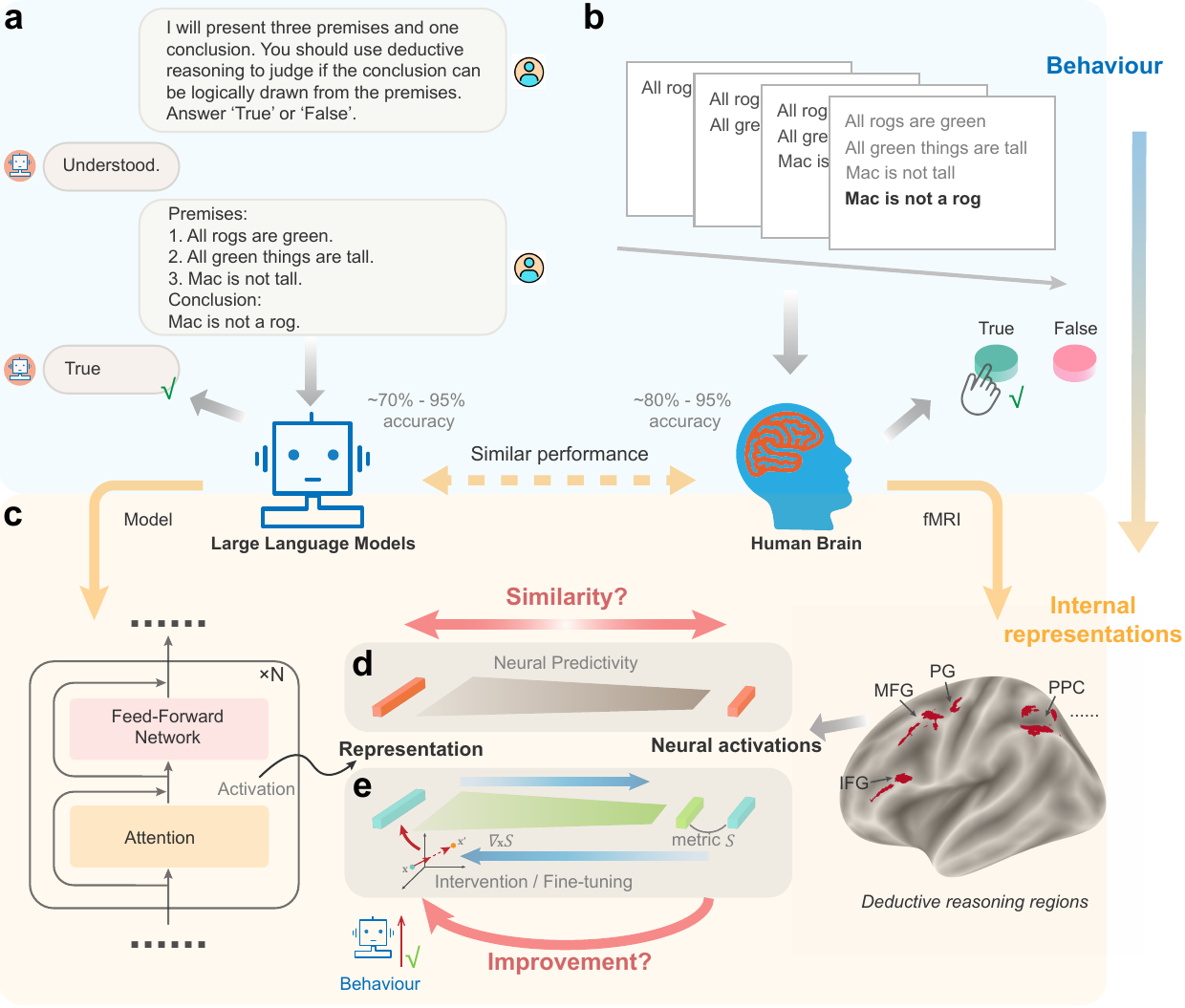}
    \caption{\textbf{Overview of the Connections between Large Language Models and Humans for Solving Deductive Reasoning Problems.} We aim to step from the behaviour level to internal representations. (a-b) \acp{llm} and humans show similar behaviour performance. (a) \acp{llm} process problems through chat-formatted prompts, generating direct linguistic responses. (b) Human participants sequentially view premises and conclusions, responding via button press after the presentation of the conclusion. (c) We examine whether shared behaviour performance reflects internal connections between internal representations of \acp{llm} and humans, \ie, if their internal representations (model representations and neural activations of human brains in related regions) align, and if neural signals can guide the improvement of \ac{ai} models. (d) We study the representational similarity through the neural predictivity score. (e) We enhance \acp{llm} by steering their representations of incorrect responses toward neural-derived directions through representation intervention or parameter fine-tuning.}
    \label{fig:overview}
\end{figure}

We consider multiple instruction-tuned \ac{llm} models as specified in the main text, and \cref{supp:sec:base-model} also presents neural predictivity results for the base Llama2-7B and Qwen2-7B models. For the instruction-tuned models, we first prompt models with the task objective and then present three premises and the conclusion sequentially (\cref{fig:overview}a; complete prompts in \cref{supp:sec:model-prompt}). Models without chain-of-thought reasoning are required to output only ``True'' or ``False'' at the beginning of the response, and we only consider this first response token; while the thinking model normally generates natural language reasoning before providing answers, and we extract final answers from the responses. We set the generation temperature as $0$, \ie, only deterministically outputting the token with the highest probability without sampling, and evaluate accuracy based on the first output token or extracted answer.

All these \ac{llm} models are based on the transformer architecture~\cite{vaswani2017attention}, which processes inputs through successive layers comprising an attention module followed by a \ac{ffn} module, except for slight differences in detailed implementations of the modules. For intermediate representations of models, we consider both attention module outputs and layer outputs of each layer for the last token. Specifically, the calculation of each layer of autoregressive transformers is:
\begin{equation}
\begin{aligned}
    \mathbf{h}_i^{l} &= \mathbf{x}_i^{l-1} + \text{Attn}^{l}\left(\mathbf{x}_{:i}^{l-1}\right), \\
    \mathbf{x}_i^{l} &= \mathbf{h}_i^{l} + \text{FFN}^{l}\left(\mathbf{h}_i^{l}\right),
\end{aligned}
\end{equation}
where $l$ and $i$ are layer and token indices, and we consider $\left\{\text{Attn}^l\left(\mathbf{x}^{l-1}_{:n}\right)\right\}_l$ and $\left\{\mathbf{x}^l_n\right\}_l$ as all representations.

\subsection{Calculation of neural predictivity}

Neural predictivity is calculated based on functionally localized model units following the latest practice~\cite{alkhamissi2025llm,alkhamissi2025language}. The calculation process involves the following steps. First, we localize model units responsive to deductive reasoning by contrasting activations between the reasoning condition and a control condition (only reading premises; \cref{supp:sec:model-prompt-control}). We choose top-$N$ most responsive units from both attention outputs and layer outputs~\cite{kumar2024shared} at the last token, where $N$ corresponds to the width of the model's layer. The activations from these units for each problem are extracted as model representations. In addition to this, we also investigate representations from each layer. 
Then, we use 80\% of the problems to fit a ridge regression between model representations and brain representations, \ie, voxel vectors, implemented via RidgeCV in the sklearn library with automatic parameter selection for ridge regression. After that, we apply the regression model to the held-out 20\% problems to generate predictions of brain representations given model representations. The predictions are compared against the corresponding brain representations with a Pearson correlation. This process is repeated five times, corresponding to a 5-fold split for train and test sets, and the predictivity score of each voxel is averaged across the five splits. Then, the raw score is calculated by taking the median of voxels and computing the median and \ac{mad} across participants. 

The raw scores are further divided by an estimated upper-bound ceiling based on predictivity between human neural activations. This process is similar to the process in~\cite{schrimpf2021neural} and can be found in \cref{supp:sec:brain-score}. The calculated ceiling values are $0.257, 0.299, 0.28$ for deductive reasoning regions considering all problems, syllogistic problems, and transitive problems; $0.248, 0.287, 0.26$ for \ac{md} regions; and $0.253, 0.293, 0.27$ for language regions. They are comparable to previous ceiling values of language regions in the language comprehension task (0.32 and 0.20 for two \ac{fmri} datasets)~\cite{schrimpf2021neural}.

\subsection{Statistical tests}

The reported predictivity scores first take the median among voxels and then compute the median and \ac{mad} (as error bars) across participants. For group-level comparisons across models, we report means with standard deviations (as error bars). Statistical testing employs two-sided paired t-tests comparing individual model scores between conditions. 
For fine-tuning experiments, statistical testing also employs two-sided paired t-tests comparing NARF+Label and Label under the same random seeds. 
All p-values less than $0.05$ are marked with a single asterisk, p-values less than $0.01$ with double asterisks, and p-values less than $0.001$ with triple asterisks.

\subsection{Neural activation guided representation intervention}\label{subsec:nari}

Our intervention approach modifies model representations during inference for incorrectly answered reasoning problems. The method proceeds through three computational stages. 

First, we establish a mapping between model representations and neural activity patterns. For each subject $sub$ and layer $l$, we fit a ridge regression (using RidgeCV with cross-validation) from the model's representations $\{\mathbf{r}^l\}$ to the corresponding fMRI voxel vectors $\{\mathbf{v}_{sub}\}$, yielding weight matrix $\mathbf{W}^l_{sub}$ and intercept $\mathbf{b}^l_{sub}$. $\mathbf{W}^l_{sub}$ is calculated based on the centered data while $\mathbf{b}^l_{sub}$ reflects the systematic shift between the mean of the source and target data. This mapping is computed separately for syllogistic and transitive reasoning problems.

Then, for incorrectly answered problems where human subjects responded correctly, we compute an intervention objective $S=\text{Sim}\left(\mathbf{W}^l_{sub}\mathbf{r}^l, \mathbf{v}_{sub}\right)$, where $\text{Sim}$ is a similarity metric (such as cosine similarity). The gradients of representations $\nabla_{\mathbf{r}^l} S$ define the direction to modify representations, which is based on the joint effect of the structures of model and human neural representations (see \cref{sup:sec:explain-narinarf} for detailed derivation). Note that we drop the intercept term when mapping the model's representations to the \ac{fmri} space during intervention direction calculation, and experiments show the improved performance of dropping this term (\cref{fig:intervention}f, \cref{sup:fig:intervention-analysis}). This means incorporating an additional systematic shift between model and human neural representations during direction calculation, apart from encouraging the similarity between centered data. The ablation results in \cref{sup:fig:intervention-analysis} further show that this systematic shift alone cannot achieve better performance, but should be combined with the structure of human neural activations, suggesting the importance of both structural alignment and systematic shifts between model and neural representations.

Finally, we optimize representations through constrained gradient ascent with these gradients. Using the Adam optimizer, we iteratively update representations while projecting modifications to remain within $\alpha \sigma^l$ distance from the original, where $\alpha$ is a specified range (\cref{fig:intervention}a) and $\sigma^l$ is the generalized standard deviation of high-dimensional representations for all reasoning problems, \ie, the root mean square of Euclidean distances from the mean. 
We perform multiple gradient ascent steps and periodically verify whether the intervened representations can correct the answer. The maximum steps are set as 200 or 50, and the verification is carried out every 5 steps. Once the correct answer is reached, the intervention terminates and is treated as successful. 

For the baseline of random signals, we replace \ac{fmri} signals with random data and carry out the same procedures. For the baseline of random directions, we perform a grid search for the direction within the restriction. All baselines with randomness are evaluated under three random seeds and report the mean and standard deviation.

For thinking models (\eg, DeepSeek-Distill-Qwen-1.5B), we compute layer-wise representations $\mathbf{r}^l$ as the mean across all thinking tokens: $\mathbf{r}^l=\frac{1}{n}\sum_{i=1}^{n}\mathbf{r}^l_{t_i}$, where $n$ denotes the number of thinking tokens. We derive the gradients $\nabla_{\mathbf{r}^l}S$ following the same steps, and uniformly apply them to all thinking tokens' representations during autoregressive generation. Specifically, given the representations $\mathbf{r}^{l'}$ after several gradient ascent steps, we calculate $\Delta \mathbf{r}^l=\mathbf{r}^{l'}-\mathbf{r}^l$ and add it to the representations to generate each thinking token.

To generalize intervention directions to novel test problems, we aggregated successful intervention directions across all training problems, computing the average direction $\overline{\Delta\mathbf{r}^l}$ for each layer. For inference, we scale these directions by a factor $\gamma$ (grid-searched in a range, \eg, $[0.1, 1.0]$ with 0.1 increments) and apply them additively to representations during forward propagation. These processes are done for syllogistic and transitive problems separately. For the baseline of random directions, we randomly sample directions for the representation intervention and perform the same grid search for the factor. 

More details of the method can be found in \cref{supp:sec:implementation-details}.

\subsection{Neural activation guided representation fine-tuning}

For the vanilla \ac{narf} without combination with language labels, we take advantage of NARI results to fine-tune models' parameters to align representations with successful intervention targets induced by neural activations. We denote the set of reasoning problems with successful intervention as $\mathcal{Q}_{train}$, whose elements are $(q, \{\Delta \mathbf{r}^{q,l}\})$ representing the reasoning question and the accumulated intervention direction for each layer. We minimize the \ac{mse} loss to encourage the representation toward the intervened point during fine-tuning: $\frac{1}{2}\left\lVert\mathbf{r}^{q,l} - \left(\mathbf{r}^{q,l}_{\text{ori}}+\Delta\mathbf{r}^{q,l}\right)\right\rVert^2_2$, where $\mathbf{r}^{q,l}$ is the representations of the current model and $\mathbf{r}^{q,l}_{\text{ori}}$ is the original representation before fine-tuning. Additionally, for originally correct problems (denoted as $\mathcal{Q}_{reg}$), we apply regularization to maintain the representation through the \ac{mse} loss as well: $\frac{1}{2}\left\lVert\mathbf{r}^{q,l} - \mathbf{r}^{q,l}_{\text{ori}}\right\rVert^2_2$. We train attention modules in the same middle layers as the intervention experiments, and the loss for each layer $l$ can be written as:
{\small
\begin{equation}
    L^l = \sum_{(q,\{\Delta \mathbf{r}^{q,l}\})\in \mathcal{Q}_{train}} \frac{1}{2}\left\lVert\mathbf{r}^{q,l} - \left(\mathbf{r}^{q,l}_{\text{ori}}+\Delta\mathbf{r}^{q,l}\right)\right\rVert^2_2 + \lambda\sum_{q\in\mathcal{Q}_{reg}} \frac{1}{2}\left\lVert\mathbf{r}^{q,l} - \mathbf{r}^{q,l}_{\text{ori}}\right\rVert^2_2,
\end{equation}
}
where $\lambda$ is a regularization coefficient. 
We utilize the Adam optimizer to update parameters for all models (learning rate $1e-6$, no weight decay). We take the best model across validation epochs and report test accuracy; both validation and testing consider all possible permutation orders of premises to avoid overfitting to the premise order. For $\mathcal{Q}_{reg}$, we take the nearest intervention point for each problem among all successful intervention by different human subjects, \ie, $\{\Delta\mathbf{r}^{q,l}\}$ with the smallest sum of norms. $\lambda$ is set as 1 for Qwen2-7B-Instruct, Llama2-7B-Chat, and Llama3-8B-Instruct, while 10 for Qwen2-1.5B-Instruct and Mistral-7B-Instruct. All models are trained for 100 epochs and validated for every 10 epochs.

\paragraph{Combination of NARF and language supervision}
When combining \ac{narf} with language labels, to enhance practical applicability, we do not build the method on the intervention results. Instead, we directly optimize the objective that induces the intervention directions, \ie, $S=\text{Sim}\left(\mathbf{W}^l_{sub}\mathbf{r}^l, \mathbf{v}_{sub}\right)$ in \cref{subsec:nari}. Specifically, we pre-calculate $\mathbf{W}^l_{sub}$ similar to \cref{subsec:nari}, and impose the objective for reasoning problems that the model answers incorrectly across all human subjects who responded correctly (denote the question set as $\mathcal{Q}$ and human subject set for each question $q$ as $\mathcal{S}^q$). 
For language supervision, we apply the \ac{ce} loss with the correct label token (``True'' or ``False'') on the last output token. Therefore, the minimization training objective can be written as:
{\small
\begin{equation}
    L = \lambda_{narf}\sum_{q\in \mathcal{Q}}\frac{1}{\vert\mathcal{S}^q\vert}\sum_{sub\in\mathcal{S}^q} \sum_l -\text{Sim}\left(\mathbf{W}^l_{sub}\mathbf{r}^{q,l}, \mathbf{v}_{sub}^q\right) + \text{CE}(o, label),
\end{equation}
}
where $\lambda_{narf}$ is a coefficient and $o$ is the output token of the model. We also take the best model across validation epochs and report test accuracy. This objective induces a task-relevant neural guidance on the subset of cases that require correction rather than maximizing a global correlation metric, and therefore is not necessarily monotonically coupled with global alignment with humans but related to sample-wise similarity (\cref{supp:sec:similarity-metrics}). 

We run each experiment 5 times under the same random seeds for statistical significance test. We apply \ac{lora}~\cite{hu2022lora} to the parameters of attention modules in the same middle layers and utilize the Adam optimizer to update them (learning rate $1e-4$, no weight decay), with accumulated gradients from all training problems. Models are trained for 100 epochs and validated every epoch (for Qwen2-72B-Instruct and Llama3.3-70B-Instruct, we train the models for 50 epochs). $\lambda_{narf}$ is set as $0.1$ for most models and $0.01$ for Qwen2-1.5B-Instruct. For the baseline of language label fine-tuning alone, it keeps the same setting and can be viewed as taking $\lambda_{narf}=0$. 

More implementation details can be found in \cref{supp:sec:implementation-details}.

\subsection*{Data Availability}
All the datasets used in the paper are publicly available or generated by the code. The fMRI dataset of deductive reasoning is available on OpenNeuro: \url{https://openneuro.org/datasets/ds003076}. The fMRI dataset of HCP relational processing is available on \url{www.humanconnectome.org}. The generated data is available in the code repository: \url{https://github.com/pkuxmq/Brain-guided_LLM}.

\subsection*{Code Availability}
The implementation code is available in \url{https://github.com/pkuxmq/Brain-guided_LLM}~\cite{xiao2026code}.

\subsection*{Acknowledgments}
Z. Lin was supported by the NSF China (No. 62276004). K. Du was supported by Tsinghua University Dushi Program (20261080143).

\subsection*{Author Contributions Statement}
Conceptualization, methodology, and experiments: M.X.; Investigation and analysis: M.X., K.D.; Supervision: Z.L., K.D.; Writing: M.X., K.D., Z.L.

\subsection*{Competing Interests Statement}
The authors declare no competing interests.

\clearpage


\clearpage

\begin{figure}[t!]
    \centering
    \includegraphics[width=\linewidth]{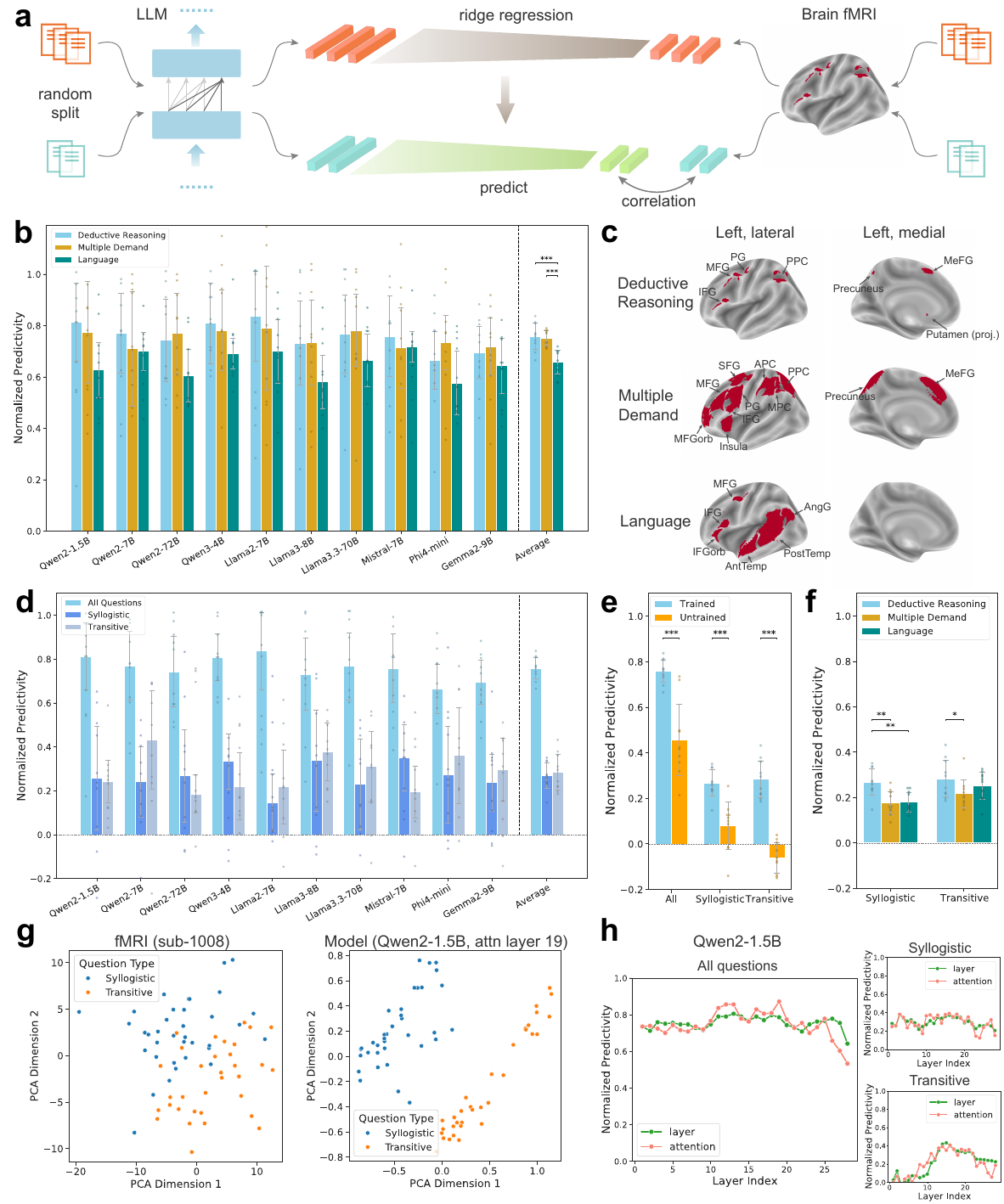}
    \caption{\textbf{Results of Neural Predictivity.} (a) Neural predictivity calculation pipeline: Fitting ridge regression to map model representations to brain activations, and computing held-out correlation scores divided by ceiling values (median across voxels and participants). (b) Neural predictivity scores of different models across brain regions (deductive reasoning, \ac{md}, language) for all reasoning problems. Data are presented as median $\pm$ \ac{mad} considering inter-subject variability for each model's results ($n=10$), and as mean $\pm$ std considering inter-model variability for the Average results ($n=10$), with individual data points overlaid (the same for (d)). Two-sided paired t-tests across models show significantly higher alignment in deductive reasoning and \ac{md} regions compared to language regions (both $p<0.001$). (c) Anatomical locations of deductive reasoning, \ac{md}, and language regions. (d) Type-specific predictivity scores (syllogistic or transitive). (e) Comparison between trained and untrained (weights randomly initialized; tokenizer trained) models. Data are presented as mean $\pm$ std considering inter-model variability ($n=10$), with individual data points overlaid (the same for (f)). Two-sided paired t-tests across models show significantly higher predictivity of trained models (all $p<0.001$). (f) Regional predictivity scores stratified by reasoning type. Two-sided paired t-tests across models are presented (for syllogistic reasoning: $p=0.001$ for Deductive Reasoning vs. MD and $p=0.009$ for Deductive Reasoning vs. Language; for transitive reasoning: $p=0.01$ for Deductive Reasoning vs. MD). (g) Representational analysis revealing type separation among model and brain representations. This separation largely accounts for the high predictivity under all questions. (h) Layer-wise predictivity scores under various settings (full results in \cref{sup:fig:different-layer}).}
    \label{fig:brain-score}
\end{figure}

\clearpage

\clearpage

\begin{figure*}[t!]
    \centering
    \includegraphics[width=\linewidth]{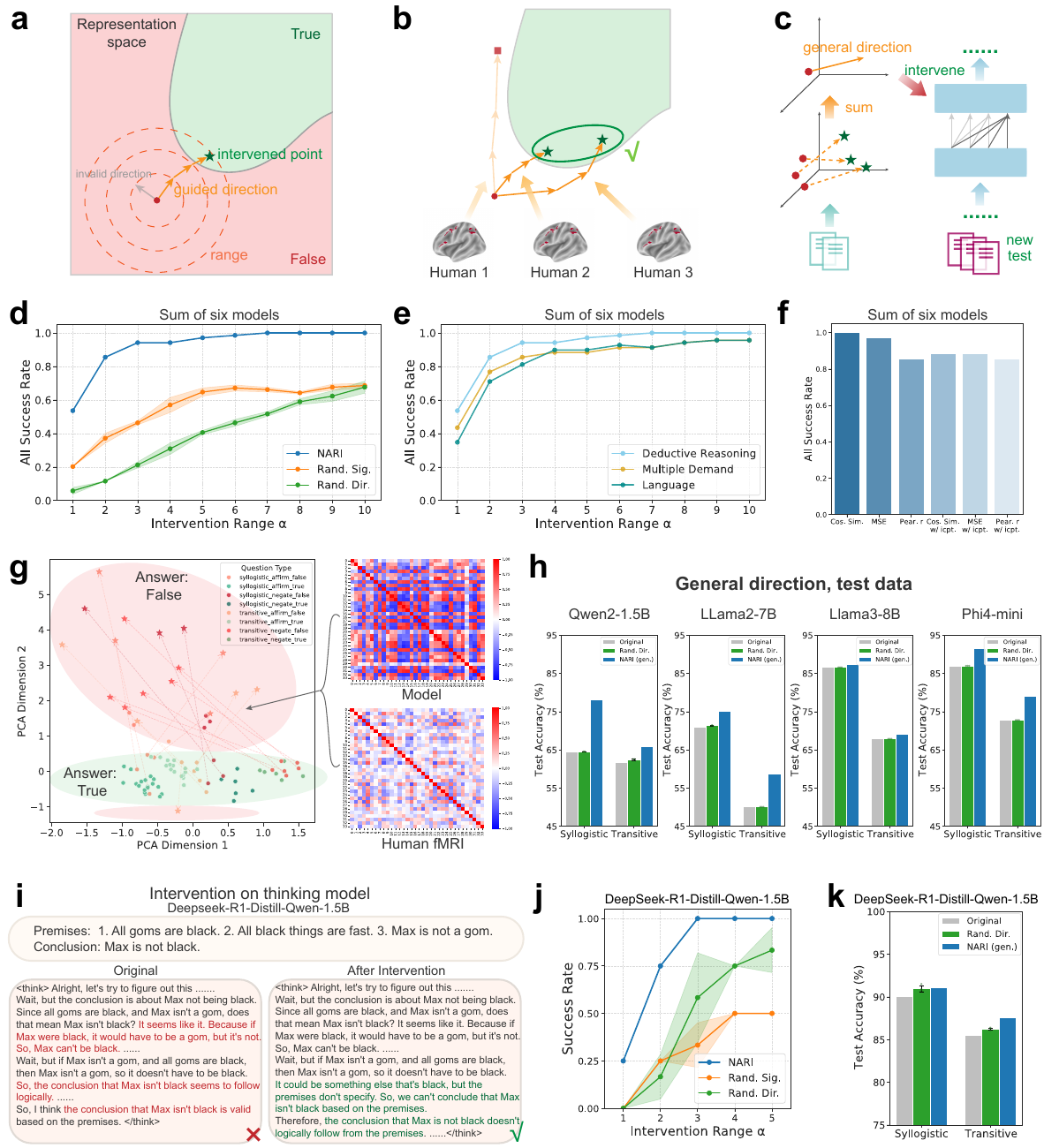}
    \caption{\textbf{Results of Representation Intervention Experiments.} (a) Illustration of the representation intervention paradigm. Effective directions in the representation space can guide the representation of incorrect responses toward correct model outputs, with smaller required modification ranges indicating higher precision of the direction. (b) Different \ac{fmri} signals from various human subjects may lead to diverse intervention results. We integrate all human subjects and consider a problem successful if a valid intervention exists. (c) We integrate successful directions into a general direction, enabling performance improvement on new test problems through inference-time intervention. (d) Summary results of six \ac{llm} models for intervention on their incorrect problems. Our \ac{nari} achieves a 100\% success rate and much higher success rates under small range restrictions than baselines of random signals (eliminate the structure of \ac{fmri} representations and rely only on model representational structures) and the max-integration of random directions. For baselines, data are presented as mean $\pm$ std over three runs (the same for (j)). (e) Comparison results of different brain regions. Deductive reasoning regions yield higher success rates than \ac{md} and language networks. (f) Comparison results of different objectives for direction calculation (\ie, gradients), including cosine similarity (Cos. Sim.), mean square error (MSE), pearson r (Pear. r), and their counterpart with intercept (icpt.). (g) Visualization of the intervention process for Qwen2-1.5B-Instruct in the representation space (attention output of the 3/4-th layer). The joint effect of the structures of model and \ac{fmri} representations leads to successful intervention directions. (h) Results of the general direction on the new test data. For Rand. Dir., data are presented as mean $\pm$ std over three runs (the same for (k)). (i) Demonstration of adapting \ac{nari} to chain-of-thought reasoning models, enabling intervention on the language-based thinking process. (j) Intervention results on incorrect problems of the thinking model DeepSeek-R1-Distill-Qwen-1.5B. (k) General direction results on the thinking model DeepSeek-R1-Distill-Qwen-1.5B.}
    \label{fig:intervention}
\end{figure*}

\clearpage

\clearpage

\clearpage
\appendix
\setcounter{section}{0}
\setcounter{figure}{0}
\setcounter{equation}{0}
\setcounter{table}{0}
\renewcommand\thesection{S\arabic{section}}
\renewcommand\thefigure{S\arabic{figure}}
\renewcommand\theequation{S\arabic{equation}}
\renewcommand\thetable{S\arabic{table}}

\markboth{Supplementary Information}{Supplementary Information}
\section{More details of the fMRI dataset and preprocessing}\label{supp:sec:fmri-preprocess}

The \ac{fmri} dataset (publicly available on OpenNeuro, ID: ds003076) is an accompanying adult dataset to a school-aged children dataset for deductive reasoning~\cite{lytle2020children} and was also part of the data used in~\cite{prado2013fractionating}. We focus on this adult dataset for its higher behaviour accuracy, enabling robust comparison with and improvement for \acp{llm}. 

The dataset involves two types of deductive reasoning: syllogistic and transitive reasoning. For syllogistic reasoning problems, the premises describe a series of relationships (adjective) among three classes (monosyllabic pseudoword), and the adjective is one of the following: \textit{tall, short, big, small, old, young, fast, slow, brown, red, black, blue, green, white, pink}. For transitive reasoning problems, the premises describe a linear ordering (comparative adjective) of 4 imaginary characters (syllable name), and the comparative adjective is one of the following: \textit{slower, faster, shorter, taller, younger, older, smaller, bigger}. 

We use MATLAB, SPM12, and GLMSingle~\cite{prince2022improving} to preprocess \ac{fmri} data. We first preprocess the data with slice time correction, motion correction (realigning all functionals to the mean image of the first functional run), normalization to the standard \ac{mni} space by first coregistration to the structural image and then normalization with estimated parameters during segmentation of the structural image (normalized voxel size $2\times 2\times 2 \, \text{mm}^3$), and smoothing with a 4mm FWHM Gaussian filter. 

Then we estimate the response to each problem with GLMSingle, which is an improved method to estimate single-trial \ac{fmri} responses based on custom \ac{hrf} for each voxel, derivation of a set of noise regressors with cross-validation, and ridge regression for beta estimates~\cite{prince2022improving}. We separate this process for syllogistic and transitive problems because their response time differs in general, and use their average response time as the response time in GLMSingle. For the design matrix, we treat each type of problems with correct or incorrect answers as a condition (used for cross-validation in GLMSingle), resulting in up to $8\times2=16$ conditions depending on each human subject's behaviour, and extract single-trial responses for each problem with the design matrix.

We extract voxels from the estimated response based on locations of deductive reasoning regions specified in the meta-analysis~\cite{prado2011brain} and a process similar to group-constrained, participant-specific functional localization~\cite{fedorenko2010new,schrimpf2021neural}. 
Specifically, we first construct a deductive reasoning region mask based on spherical approximations of the coordinates and volume sizes (minimum radius set as 5\,mm) of peaks for all deductive arguments, relational arguments (\ie, transitive), and categorical arguments (\ie, syllogistic) specified in \cite{prado2011brain} (Table 2 and Table 3, we transform the provided Talairach coordinates into the \ac{mni} space), yielding a mask involving left \ac{ifg}, left \ac{mefg}, bilateral \ac{mfg}, bilateral \ac{pg}, bilateral \ac{ppc}, and left \ac{bg}. 
Then we compute individual activation maps within this mask using a contrast condition. Specifically, for each human subject, we first use SPM12 to get the t-values of all voxels considering the contrast: correct\_reasoning $>$ premises, where the condition correct\_reasoning refers to correctly answered problems (onset time: the start of judgement after presenting the conclusion, duration: response time to the button press), while the condition premises refers to reading premises (onset time: the start of presentation of premises, duration: 6s). The t-values are separately calculated for syllogistic and transitive problems, and we take the max of them for each voxel to select 10\% of the most responsive voxels within the mask (with the highest t-values, responsive to either syllogistic or transitive reasoning). These voxels are considered as vectors of brain representations for each human subject. For language and \ac{md} regions, their masks are available from \url{https://www.evlab.mit.edu/resources}, which are widely used in previous works~\cite{schrimpf2021neural}, and we follow the same procedure as deductive reasoning regions.

\section{More details of the test datasets}

\subsection{Generated reasoning problems}\label{supp:sec:generated-data}

\paragraph{Syllogistic and transitive problems} 
The generated questions follow the structure of questions in the \ac{fmri} dataset, while introducing new lexical items and premise numbers. The monosyllabic pseudowords and syllable names are randomly generated and different from the \ac{fmri} dataset, while the adjectives are randomly chosen from a list -- for syllogistic reasoning, the list of adjectives is: \textit{strong, weak, bright, dark, warm, cold, soft, hard, smooth, rough, quiet, loud, sharp, dull, heavy, light, wide, narrow, deep, shallow, clean, dirty, fresh, stale, rich, poor, brave, cowardly, cheerful, sad, proud, humble}; for transitive reasoning, the list of comparative adjectives is: \textit{brighter, darker, richer, poorer, stronger, weaker, quieter, louder, cleaner, dirtier, fresher, staler, heavier, lighter, smoother, rougher, warmer, colder, wider, narrower, deeper, shallower, sharper, duller, softer, harder, prouder, humbler, braver, calmer, sadder, happier}, all distinct from the original \ac{fmri} stimuli. 

For three-premise problems, we maintain the original \ac{fmri} dataset's eight conditions (true/false $\times$ affirmative/negation $\times$ 2/3 required premises). For problems with more premises (4-6), we consider all premises to be used and organize problems into four conditions (true/false $\times$ affirmative/negation). We generate the same number of questions for all groups, enabling a more balanced measurement than the \ac{fmri} dataset. 

During validation and testing, we enumerate all possible premise permutations ($N!$ variants for $N$ premises). For example, for three premises (1) \textit{A is stronger than B}, (2) \textit{B is stronger than C}, (3) \textit{C is stronger than D}, there will be six possible orders, such as (1) \textit{C is stronger than A}, (2) \textit{B is stronger than C}, (3) \textit{A is stronger than B}. This controls the overfitting to the fixed pattern of linear orders and comprehensively evaluates the deductive reasoning ability of models that may differ for different premise orders~\cite{chen2024premise}.

For the validation set, we consider problems with three premises and generate 10 problems for each condition, resulting in $10\times8\times2=160$ problems and finally $160\times3!=960$ validation data. For the test set with three/four/five/six premises, we generate 100/50/20/10 problems for each group, leading to 1600/400/160/80 problems and finally 9600/9600/19200/57600 test data.

\paragraph{Propositional problems}
We generate propositional problems with three types: modus ponens, modus tollens, and disjunction elimination~\cite{prado2011brain}, each with true or false conclusions. For modus ponens with the correct conclusion, the two premises are in the format: (1) \textit{If pseu1 is adj1, then pseu2 is adj2}, (2) \textit{pseu1 is adj1}, and the conclusion is \textit{pseu2 is adj2}, where \textit{pseu1} and \textit{pseu2} are two randomly generated pseudowords and \textit{adj1} and \textit{adj2} are two adjectives randomly chosen from: \textit{strong, weak, bright, dark, warm, cold, soft, hard, smooth, rough, quiet, loud, sharp, dull, heavy, light, wide, narrow, deep, shallow, clean, dirty, fresh, stale, rich, poor, brave, cowardly, cheerful, sad, proud, humble}. For modus ponens with the incorrect conclusion, the conclusion is \textit{pseu2 is not adj2}. For modus tollens, the second premise is replaced by \textit{pseu2 is not adj2} and the conclusion is \textit{pseu1 is not adj1} or \textit{pseu1 is adj1} correspondingly. For disjunction elimination, the two premises are in the format: (1) \textit{pseu1 is adj1 or pseu2 is adj2}, (2) \textit{pseu1 is not adj1}, and the conclusion is \textit{pseu2 is adj2} or \textit{pseu2 is not adj2}. We generate 100 problems for each condition, resulting in 600 problems.

\subsection{FOLIO dataset}\label{supp:sec:folio}

The FOLIO dataset~\cite{han2024folio} is a first-order logical inference dataset for reasoning in natural language. Each question consists of multiple premises and a hypothesis. An example of the dataset is:

Premises:

(1) The Blake McFall Company Building is a commercial warehouse listed on the National Register of Historic Places.

(2) The Blake McFall Company Building was added to the National Register of Historic Places in 1990.

(3) The Emmet Building is a five-story building in Portland, Oregon.

(4) The Emmet Building was built in 1915.

(5) The Emmet Building is another name for the Blake McFall Company Building.

(6) John works at the Emmet Building.

Hypothesis:

A five-story building is built in 1915.

We follow the FOLIO-wiki-curated version~\cite{zhang2023cumulative}, a refined subset removing problematic instances. The answers to the hypothesis can be ``True'', ``False'', or ``Unknown'', and we only consider questions with ``True'' or ``False'' answers (\ie, FOLIO-wiki-curated-TF) in order to align with the setting of our training data, with a total number of 315 questions.

\section{More details of model prompting}\label{supp:sec:model-prompt}

The prompt for syllogistic reasoning is:

``\textit{You are an expert in performing logical reasoning. I will present three premises and one conclusion each round. The premises describe a series of relationships among monosyllabic pseudowords and adjectives. Assume all premises are True. You should use deductive reasoning to determine if the conclusion can be drawn from the premises logically. Answer `True' if the conclusion can be drawn logically; otherwise you should answer `False'. Only answer `True' or `False' at the beginning of the response.}'' 

For transitive reasoning, the description is slightly different: 

``\textit{The premises describe relationships among imaginary characters with comparative adjectives.}'' 

For problems with more premises, the description correspondingly changes to the number of premises. And for propositional reasoning, the corresponding description is:

``\textit{I will present two premises and one conclusion each round. The premises describe logical arguments.}'' 

After the model's round saying 
``\textit{Understood. I will do my best to determine if the conclusion can be logically drawn from the given premises, and only answer `True' or `False' at the beginning of the response.}'', 
the premises and conclusion are presented, \eg, 
``\textit{Premises:$\backslash$n 1. All blons are pink.$\backslash$n 2. All pink things are young.$\backslash$n 3. Ken is a blon.$\backslash$n Conclusion:$\backslash$n Ken is pink.}'' 
Then the model is required to output the answer at its round in the chat template.

\subsection{Prompt for the control condition}\label{supp:sec:model-prompt-control}

To functionally localize model units responsive to deductive reasoning for neural predictivity calculation, we further design a control condition to only read premises. Under this condition, the task prompt of syllogistic premises to the model is:

``\textit{I will present three premises that describe a series of relationships among monosyllabic pseudowords and adjectives. You should just read the descriptions and do nothing. Only respond `I have read the premises' after my messages.}''

For transitive premises, the description is also slightly different as above. After the model's round saying ``\textit{Understood.}'', only the three premises are presented.

\section{More details of ceiling calculation for neural predictivity}\label{supp:sec:brain-score}

For neural predictivity calculation, the raw scores are divided by an estimated upper-bound ceiling based on predictivity between human neural activations. This is due to the intrinsic noise in biological measurements, so a ceiling value reflecting the explainable variance in neural activations is usually adopted for brain scores. We calculate the ceiling value in a similar way as the established practice~\cite{schrimpf2021neural}. First, we subsample the $n$-participant data into all possible combinations of $s$ participants ($s\in [2, n]$). For each subsample, we designate a target participant in the subsample and predict its neural signals from the remaining participants using ridge regression that matches the main analysis pipeline. Then we calculate the highest possible estimate for each voxel of each human subject by extrapolating the score to infinitely many participants through fitting the equation $v=v_0\times \left(1 - e^{-\frac{s}{\tau_0}}\right)$ for each voxel, where $v$ represents the score and $v_0$ and $\tau_0$ are parameters to fit. Finally, the ceiling score takes the average of $v_0$ for all voxels of all human subjects. The calculated ceiling values are 0.257, 0.299, 0.28 for deductive reasoning regions considering all problems, syllogistic problems, and transitive problems; 0.248, 0.287, 0.26 for \ac{md} regions; and 0.253, 0.293, 0.27 for language regions. They are comparable to previous ceiling values of language regions in the language comprehension task (0.32 and 0.20 for two \ac{fmri} datasets)~\cite{schrimpf2021neural}.

\section{More explanations of NARI and NARF}\label{sup:sec:explain-narinarf}

Both neural activation guided representation intervention (NARI) and fine-tuning (NARF) leverage the joint structures of model representations and human neural activations. Specifically, consider the ridge regression from the space of model representations to the space of neural signals, let $\mathbf{X}$ and $\mathbf{Y}$ denote the data of model representations and neural activations (columns as samples), the weight $\mathbf{W}$ and intercept $\mathbf{b}$ of ridge regression ideally approximate the closed-form solution:
\begin{equation}
    \begin{aligned}
        \mathbf{W} &= \mathbf{Y}_c^\top\mathbf{X}_c\left({\mathbf{X}_c^\top\mathbf{X}_c+\lambda\mathbf{I}}\right)^{-1}, \\
        \mathbf{b} &= \overline{\mathbf{y}} - \mathbf{W}\overline{\mathbf{x}},
    \end{aligned}
\end{equation}
where $\overline{\mathbf{x}}$ and $\overline{\mathbf{y}}$ are sample means, $\mathbf{X}_c=\mathbf{X}-\mathbf{1}\overline{\mathbf{x}}^\top$ and $\mathbf{Y}_c=\mathbf{Y}-\mathbf{1}\overline{\mathbf{y}}^\top$ are centered data, and $\lambda$ is the ridge coefficient.

For the objective $S=\text{Sim}(\mathbf{W}\mathbf{x},\mathbf{y})$ with the data point $\mathbf{x}$ and $\mathbf{y}$, when $\text{Sim}$ is taken as cosine similarity or the negative of \ac{mse}, the gradient of $\mathbf{x}$ is:
\begin{equation}
    \begin{aligned}
        \nabla_{\mathbf{x}}S_{cos} &= \frac{1}{\lVert \mathbf{W}\mathbf{x}\rVert\lVert y\rVert}\mathbf{W}^\top\mathbf{y}-\frac{\left<\mathbf{W}\mathbf{x}, \mathbf{y}\right>}{\lVert\mathbf{W}\mathbf{x}\rVert^3\lVert\mathbf{y}\rVert}\mathbf{W}^\top\mathbf{W}\mathbf{x},\\
        &=\frac{1}{\lVert \mathbf{W}\mathbf{x}\rVert\lVert y\rVert}\left({\mathbf{X}_c^\top\mathbf{X}_c+\lambda\mathbf{I}}\right)^{-1}\mathbf{X}_c^\top\mathbf{Y}_c\mathbf{y}\\
        &\quad -\frac{\left<\mathbf{W}\mathbf{x}, \mathbf{y}\right>}{\lVert\mathbf{W}\mathbf{x}\rVert^3\lVert\mathbf{y}\rVert}\left({\mathbf{X}_c^\top\mathbf{X}_c+\lambda\mathbf{I}}\right)^{-1}\mathbf{X}_c^\top\mathbf{Y}_c\mathbf{Y}_c^\top\mathbf{X}_c\left({\mathbf{X}_c^\top\mathbf{X}_c+\lambda\mathbf{I}}\right)^{-1}\mathbf{x},\\
        \nabla_{\mathbf{x}}S_{-mse} &= -\mathbf{W}^\top\mathbf{W}\mathbf{x}+\mathbf{W}^\top\mathbf{y}\\
        &=-\left({\mathbf{X}_c^\top\mathbf{X}_c+\lambda\mathbf{I}}\right)^{-1}\mathbf{X}_c^\top\mathbf{Y}_c\mathbf{Y}_c^\top\mathbf{X}_c\left({\mathbf{X}_c^\top\mathbf{X}_c+\lambda\mathbf{I}}\right)^{-1}\mathbf{x}\\
        &\quad +\left({\mathbf{X}_c^\top\mathbf{X}_c+\lambda\mathbf{I}}\right)^{-1}\mathbf{X}_c^\top\mathbf{Y}_c\mathbf{y}.
    \end{aligned}
    \label{sup:eq:grad}
\end{equation}

\cref{sup:eq:grad} shows that the direction (gradient) will be induced by $\mathbf{X}_c^\top\mathbf{X}_c$, $\mathbf{Y}_c\mathbf{Y}_c^\top$, and $\mathbf{X}_c^\top\mathbf{Y}_c$, which are the structure of model representations, the structure of human neural activations, and the interactive structure of model and neural representations over all considered reasoning problems. This is a joint effect of model and neural representations, which varies among different models and human subjects (\cref{supp:sec:rep-structure}).
For the baseline that replaces the \ac{fmri} data with random signals, it eliminates the effect of human neural activations ($\mathbf{Y}_c\mathbf{Y}_c^\top$ and $\mathbf{Y}_c\mathbf{y}$ are expected to be a constant matrix/vector), isolating the effect of model representational structure alone. Therefore, this ablation confirms the distinct contribution of neural guidance to representation improvement.

Additionally, this indicates that our neural-guided direction is, to some extent, robust to noises in neural signals for single voxel/time point, because it is induced by the joint structure of model and brain representations. The noise in neural representations can be mitigated in $\mathbf{Y}_c\mathbf{Y}_c^\top$ under expectation, while the remaining shared structure across problems can still interact with model representation to induce effective guidance directions.

Note that we calculate the gradients using the objective $S=\text{Sim}(\mathbf{W}\mathbf{x},\mathbf{y})$ rather than the intercept-inclusive form $S_{w/\ icpt}=\text{Sim}(\mathbf{W}\mathbf{x}+\mathbf{b},\mathbf{y})$, which empirical results demonstrate yields superior performance (\cref{supp:sec:analysis-nari}). Different from the intercept-inclusive form that aims to directly encourage the similarity between centered model and human representations, this design induces an additional term $\mathbf{W}^\top\mathbf{b}$, representing a systematic shift between model and human representations (\ie, $\mathbf{b}$) adjusted by the interactive structure between model and human representations (\ie, $\mathbf{W}$): for example, taking $\text{Sim}$ as the negative of MSE, the gradient for the intercept-inclusive form is $\nabla_{\mathbf{x}}S_{w/\ icpt}=-\mathbf{W}^\top\mathbf{W}\mathbf{x}-\mathbf{W}^\top\mathbf{b}+\mathbf{W}^\top\mathbf{y}$, while the gradient for the intercept-exclusive form is $\nabla_{\mathbf{x}}S=-\mathbf{W}^\top\mathbf{W}\mathbf{x}+\mathbf{W}^\top\mathbf{y}$, meaning $\nabla_{\mathbf{x}}S-\nabla_{\mathbf{x}}S_{w/\ icpt}=\mathbf{W}^\top\mathbf{b}$ so that the gradient ascent direction additionally incorporates a direction induced by the systematic shift. As shown in \cref{supp:sec:analysis-nari}, neither component alone (systematic shift alone or without this term) achieves optimal results. This indicates that not only structures from centered representations are important but also systematic shift between mean representations can contribute to guiding directions when modulated by $\mathbf{W}$ from the joint structures. Our method thus provides a new way to induce directions for improving model representations, instead of directly maximizing similarity scores that usually measure the similarity for centered data and have been shown not to guarantee encoding task-relevant information~\cite{cloos2025differentiable}.

\section{More implementation details of NARI and NARF}\label{supp:sec:implementation-details}

In experiments, $\text{Sim}$ is taken as cosine similarity by default, which empirically shows slightly better performance (\cref{fig:intervention}f). Overall, cosine similarity is similar to the negative of \ac{mse} because they produce similar gradient components (\cref{sup:eq:grad}). For the combination of \ac{narf} and language supervision, we take $\text{Sim}$ as the negative of \ac{mse} for Qwen2-7B-Instruct and Llama2-7B-Chat to achieve slightly higher performance. In practice, to align with the common gradient descent framework, we minimize $1-\text{CosSim}(x, y)$ or $\text{MSE}(x, y)$ and apply gradient descent.

For \ac{nari}, we apply projected gradient descent to modify representations based on the objective. Specifically, given the gradient $\nabla_{\mathbf{r}^l}S$, we first perform gradient descent with the Adam optimizer to obtain the updated representation ${\mathbf{r}^l}'$ and the difference $\Delta \mathbf{r}^l={\mathbf{r}^l}'-\mathbf{r}^l$, and then project the difference in the specified range $\alpha \sigma^l$, \ie, $\Delta \mathbf{r}^l_p=\frac{\min\left(\lVert\Delta \mathbf{r}^l\rVert, \alpha \sigma^l\right)}{\lVert\Delta \mathbf{r}^l\rVert}\Delta \mathbf{r}^l$, to get the projected representation ${\mathbf{r}^l}'_p=\mathbf{r}^l+\Delta \mathbf{r}^l_p$. We perform multiple gradient descent steps and verify if the modification direction enables the model to correct its answer for every 5 steps. The verification performs intervention for representations of all considered layers, \ie, the representation of each layer will add the difference $\Delta \mathbf{r}^l_p$ during forward propagation, and the modification can be accumulated across layers. Once the intervention corrects the answer, we stop the steps and treat it as successful. The maximum gradient descent steps are set as 200 for non-thinking models and 50 for the thinking model DeepSeek-Distill-Qwen-1.5B. The successfully intervened representation will be saved for the vanilla \ac{narf} method. For the baseline of random signals, it replaces \ac{fmri} data with randomly generated data and follows the same process. For the baseline of random directions, we sample multiple directions equivalent to the number of directions of \ac{nari}, perform grid search for the scale of the direction from $\sigma^l$ to $\alpha \sigma^l$, and treat the intervention as successful as long as the answer can be corrected during grid search.

For the general direction of \ac{nari}, we take the average of successful intervention directions $\Delta \mathbf{r}^l_p$ for all questions, with $\alpha=10$ for non-thinking models and $\alpha=5$ for DeepSeek-Distill-Qwen-1.5B. For DeepSeek-Distill-Qwen-1.5B, we only consider the nearest successful intervention point for each question. The scale $\gamma$ is searched between 0.1 and 1.0 with the interval 0.1 for Qwen2-1.5B-Instruct, Llama2-7B-Chat, and Phi-4-mini-Instruct, between 0.02 and 0.2 with the interval 0.02 for Llama3-8B-Instruct, and between 0.5 and 5.0 with the interval 0.5 for DeepSeek-Distill-Qwen-1.5B.

Considering computational costs, for the general direction of NARI (gen.), the intervention is a simple additive steering vector applied to intermediate representations in middle layers using precomputed directions. This introduces negligible overhead beyond the standard forward pass, and is feasible for real-time applications like other steering methods~\cite{li2023inference}. For the per-instance direction calculation of NARI, the gradient ascent for representations only has small costs because it mainly involves one linear transformation for the objective calculation, while the verification process requires inferring the whole model with intervention directions. Therefore, the costs mainly come from the verification process, which can take between 1-40 (or 1-10) times of LLM inference time for each fMRI data depending on the difficulty of successful intervention.

We primarily followed the prior practice of inference-time intervention~\cite{li2023inference,zhulanguage} to perform intervention on attention outputs. It is also possible to intervene in other components, such as the layer outputs or FFN outputs. We test intervention on layer output and we can similarly achieve 100\% intervention success rate together with similar improvements for the general direction setting. This suggests that our method is not limited to the attention output.

For the combination of \ac{narf} and language labels, we start training from the original model by default. For Llama2-7B-Chat, we find that starting from the \ac{narf}-trained model leads to higher performance and report this result. We do not add the regularization term in this setting. For Qwen3-4B, Gemma2-9B-it, and Llama3.3-70B-Instruct, we additionally filter three human subjects whose overall success rates for the intervention experiments are below 60\% (see \cref{sup:fig:intervention-analysis}c), and use the remaining human neural signals for experiments. We run each experiment for 5 times with the same random seeds: 0, 1, 42, 1000, 2024, and report statistical significance based on two-sided paired t-test among the 5 runs.

\section{More neural predictivity results and discussions}\label{supp:sec:diff-layer}

\begin{figure}[t!]
    \centering
    \includegraphics[width=\linewidth]{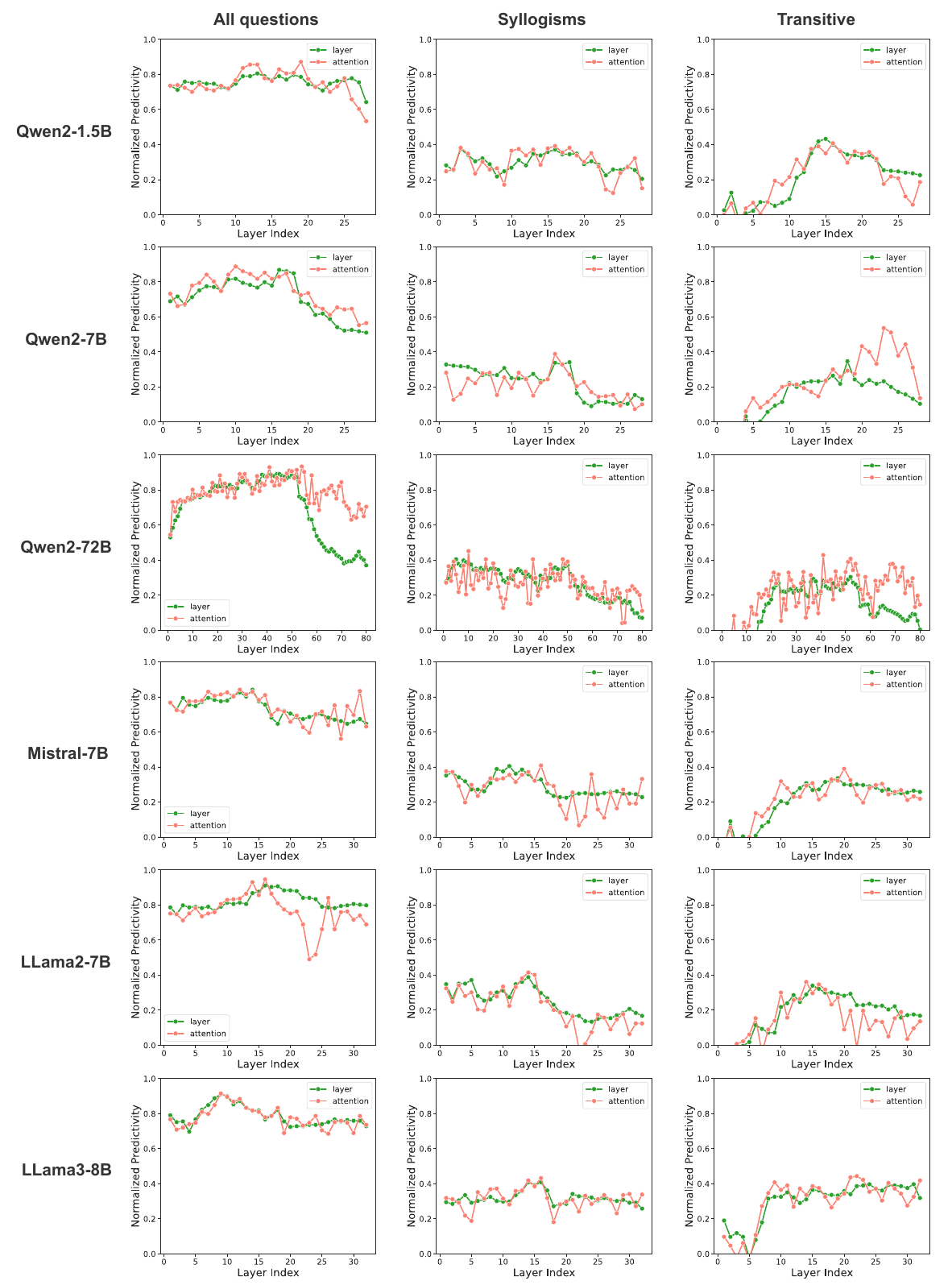}
    \caption[Layer-wise Neural Predictivity Results of Different Models]{\textbf{Layer-wise Neural Predictivity Results of Different Models.} For all models, the middle layers generally have higher scores, and attention module outputs generally have slightly higher scores than layer outputs. Models are all instruction-tuned by default, with their names shortened in the figures.}
    \label{sup:fig:different-layer}
\end{figure}

\paragraph{Layer-wise neural predictivity}
\cref{sup:fig:different-layer} presents the predictivity scores of different layers and modules (attention module output or layer output) for various models. Results show that the middle layers are generally more predictive, which may accord with the common practice to analyze middle layers because they are likely to contain interesting and abstract features~\cite{templeton2024scaling}. 
Results also show that the attention module outputs usually have slightly higher scores than layer outputs, aligning with \cite{kumar2024shared} that emphasize the importance of analyzing ``transformations'' (\ie, the output of attention heads).

\paragraph{Discussion on results of untrained models}
We note that while untrained models have near zero neural predictivity for each type of reasoning, they have above-zero predictivity scores under the setting of all reasoning questions (\cref{fig:brain-score}e). This is likely driven by the separation between reasoning types. As shown in \cref{fig:brain-score}g, brain representations exhibit certain type separation. At the same time, the untrained models in our analysis retain the trained tokenizer, so coarse linguistic differences between descriptions of syllogistic and transitive questions may still induce a corresponding separation in model representations. This potential confounder leads to above-zero predictivity score of untrained models. Despite this, trained models have much larger scores than untrained models, indicating that models capture brain representations beyond these confounding and noisy causes.

\section{Comparison of neural predictivity results between instruction-tuned and base models}\label{supp:sec:base-model}

\begin{figure}[t!]
    \centering
    \includegraphics[width=\linewidth]{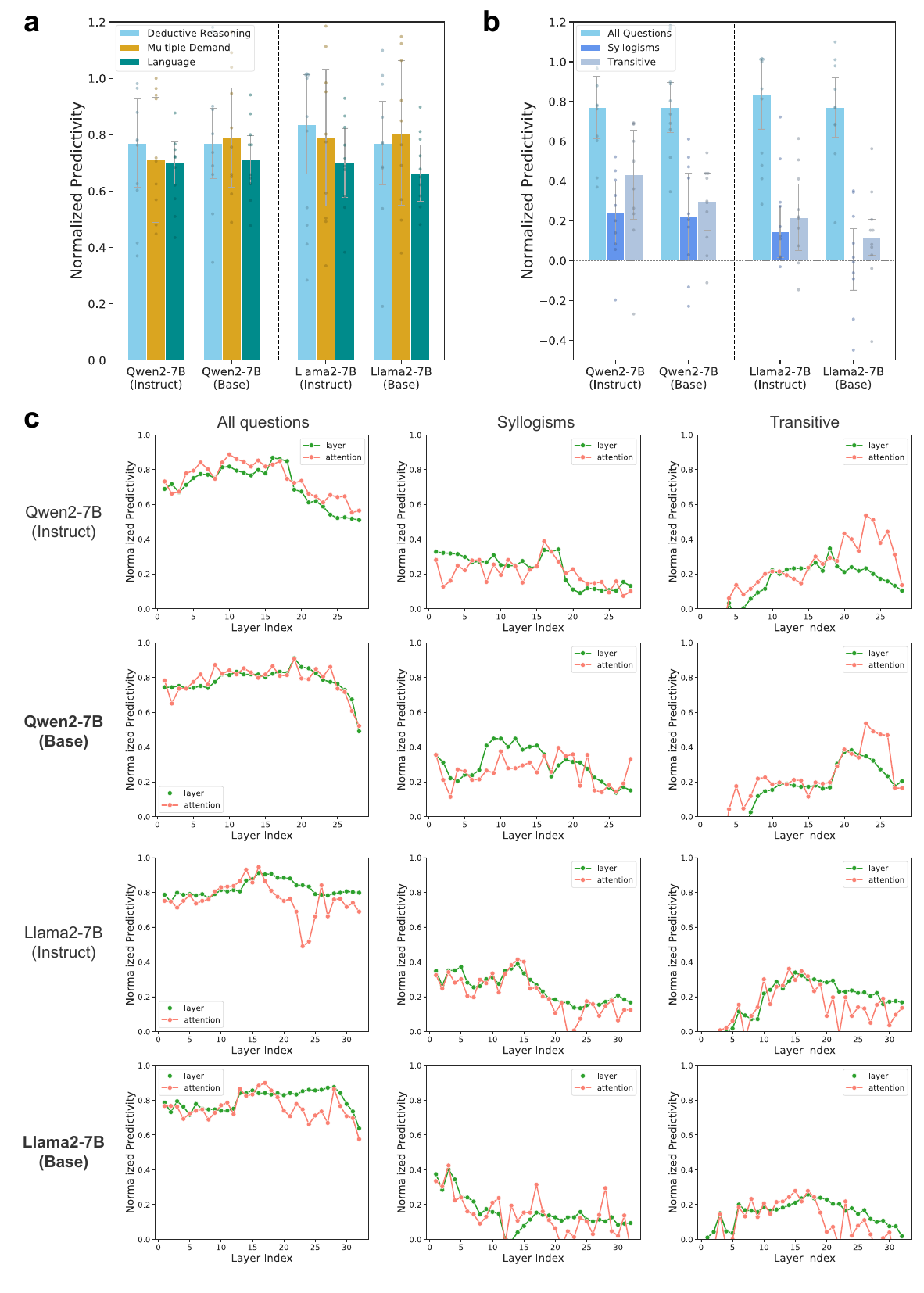}
    \caption[Comparison Results of Neural Predictivity between the Instruction-tuned and Base Models]{\textbf{Comparison Results of Neural Predictivity between the Instruction-tuned and Base Models.} (a) Neural predictivity results for different brain regions considering all reasoning problems. Data are presented as median $\pm$ \ac{mad} considering inter-subject variability for each model's results ($n=10$), with individual data points overlaid (the same for (b)). (b) Neural predictivity results for different reasoning types. (c) Layer-wise neural predictivity results.}
    \label{sup:fig:base-model}
\end{figure}

We compare the neural predictivity results between Qwen2-7B-Instruct and Qwen2-7B (base model) as well as Llama2-7B-Instruct and Llama2-7B (base model) in \cref{sup:fig:base-model}. For the base model without chat format, we present the premises and conclusion directly after the task prompt and add ``Answer:$\backslash$n'' as the generation prompt. Representations are also extracted at the last token. Results show that the base models are similar to the instruction-tuned models, considering the comparable predictivity scores, the same specificity over brain regions, and the similar discrepancy of predictivity under various reasoning type settings. 
The layer-wise score distributions also show similar trends.

\section{Structures of models' and human brains' representations}\label{supp:sec:rep-structure}

\begin{figure}[t!]
    \centering
    \includegraphics[width=\linewidth]{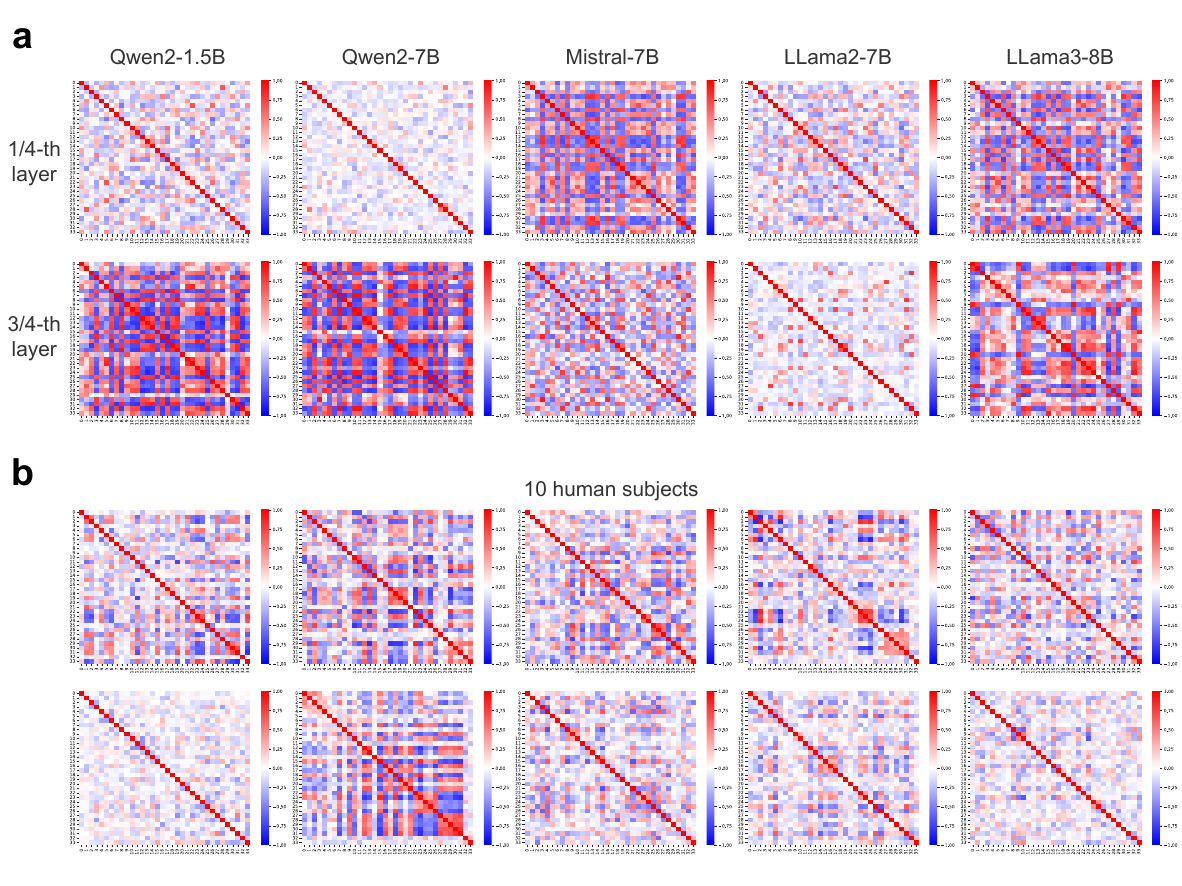}
    \caption[Visualization of the Structures of Model Representations and Human Neural Activations]{\textbf{Visualization of the Structures of Model Representations and Human Neural Activations.} The structures are the Gram matrices of centered representations for 34 transitive reasoning problems. (a) The structures of different models' representations at the 1\/4-th layer and 3\/4-th layer, showing model-dependent and layer-dependent variations. (b) The structures of neural representations from 10 human subjects, demonstrating individual-specific variations.}
    \label{sup:fig:model-fmri-structure}
\end{figure}

As analyzed in \cref{sup:sec:explain-narinarf}, the proposed \ac{nari} and \ac{narf} are induced jointly by the structures of model and human neural representations. We visualize such structures (the Gram matrices $\mathbf{X}_c^\top\mathbf{X}_c$ and $\mathbf{Y}_c\mathbf{Y}_c^\top$ of centered data as described in \cref{sup:sec:explain-narinarf}) for different models and human subjects in \cref{sup:fig:model-fmri-structure}, considering 34 transitive reasoning problems. Results show that the structures vary among different models, layers of the same models, and different human subjects. Therefore, the joint effect of them in \ac{nari}/\ac{narf} would be sample-specific, motivating our multi-subject integration approach to capture robust improvement signals.

\section{Extended NARI results}

\subsection{Model-specific representation intervention results}\label{supp:sec:intervention-each-model}

\begin{figure}[t!]
    \centering
    \includegraphics[width=\linewidth]{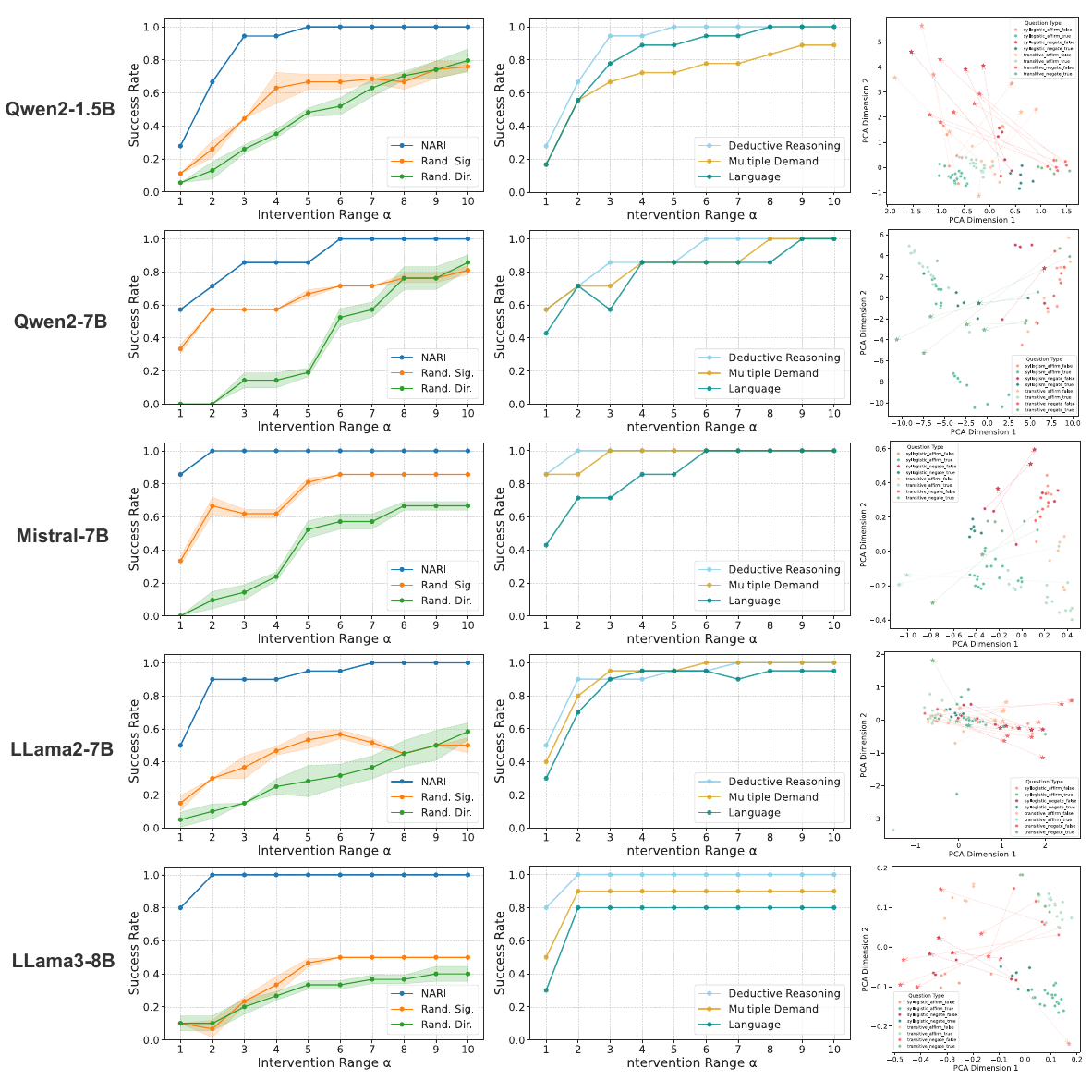}
    \caption[Model-specific Representation Intervention Results]{\textbf{Model-specific Representation Intervention Results.} The first column contains comparison results with the baselines of random signals and random directions. For baselines, data are presented as mean $\pm$ std over three runs. The second column contains comparison results of different brain regions. The third column presents visualizations of the intervention processes.}
    \label{sup:fig:intervention-each-model}
\end{figure}

The reported results of \ac{nari} in the main text (\cref{fig:intervention}) are the summary of six models. We further report the detailed results of individual models in \cref{sup:fig:intervention-each-model}, showing that the conclusion is consistent among all models.

\subsection{More analysis results of NARI}\label{supp:sec:analysis-nari}

In this section, we provide more analysis results to show the importance of incorporating the systematic shift term in \ac{nari} as discussed in \cref{sup:sec:explain-narinarf} and the variance among human subjects. 

\begin{figure}[t!]
    \centering
    \includegraphics[width=\linewidth]{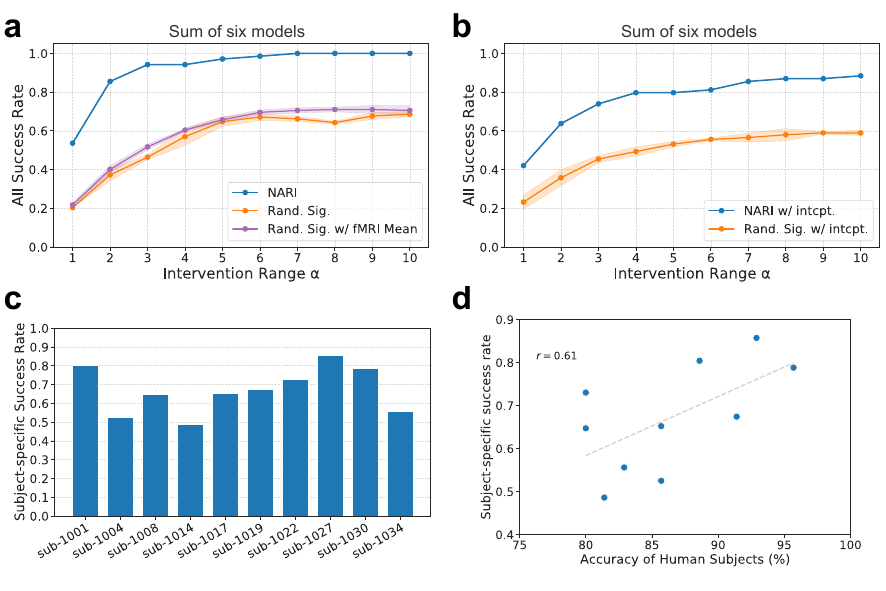}
    \caption[More Analysis Results of Representation Intervention Experiments]{\textbf{More Analysis Results of Representation Intervention Experiments.} (a) Results with the additional systematic shift term during direction calculation. ``Rand. Sig. w/ fMRI Mean'' refers to replacing \ac{fmri} signals with random signals whose mean vector keeps the same as \ac{fmri} signals. For baselines, data are presented as mean $\pm$ std over three runs (the same for (b)). (b) Results without the additional systematic shift term, \ie, calculating similarity metrics with the intercept term. (c) Subject-specific success rate of different human subjects. It is also the sum of five models. (d) The relation between subject-specific success rate and the accuracy of human subjects.}
    \label{sup:fig:intervention-analysis}
\end{figure}

First, the systematic shift term proves essential for effective representation steering. \cref{sup:fig:intervention-analysis}a-b demonstrate its superiority across all intervention ranges, with three key observations: 1) Random signal baselines also show improved success rates when incorporating the shift term (which is a shift between transformed model representations and zero), indicating its role in correcting model representation biases; 2) The shift term alone (Rand. Sig. w/ fMRI Mean condition in \cref{sup:fig:intervention-analysis}a) yields negligible effects, confirming that neural signal structure remains indispensable; 3) optimal performance requires the joint effect between neural structure and systematic shift terms. 

Second, human subject variability impacts intervention outcomes. \cref{sup:fig:intervention-analysis}c reveals substantial variation in subject-specific success rates of six models (range: 52-86\%), where the subject-specific success rate is computed as the percentage of successfully intervened reasoning problems among all problems where the model answers incorrectly while the particular human subject is correct (the problems are different for distinct human subjects). The variability aligns with the representational diversity shown in \cref{supp:sec:rep-structure}, indicating that the joint effect of structures for intervention is sample-specific. 

Considering the sum of six models, the subject-specific success rate shows quite strong positive correlation with individual human accuracy across the 10 subjects ($r=0.61$, $p=0.059$, \cref{sup:fig:intervention-analysis}d), suggesting that higher-performing subjects can not only provide more guiding problems (they answered correctly) for intervention but also induce more effective neural-guided directions for model representations. While we also find that this correlation is sensitive to model families, where Llama2-7B and Llama3-8B contribute most to the overall correlation ($r=0.68$ for Llama2-7B, $r=0.50$ for Llama3-8B, and $r=0.09$ for the sum of remaining four models). These results suggest a potential while sensitive correlation.

\subsection{Analysis results of NARI (gen.) over intervention scale}

\begin{figure}[t!]
    \centering
    \includegraphics[width=\linewidth]{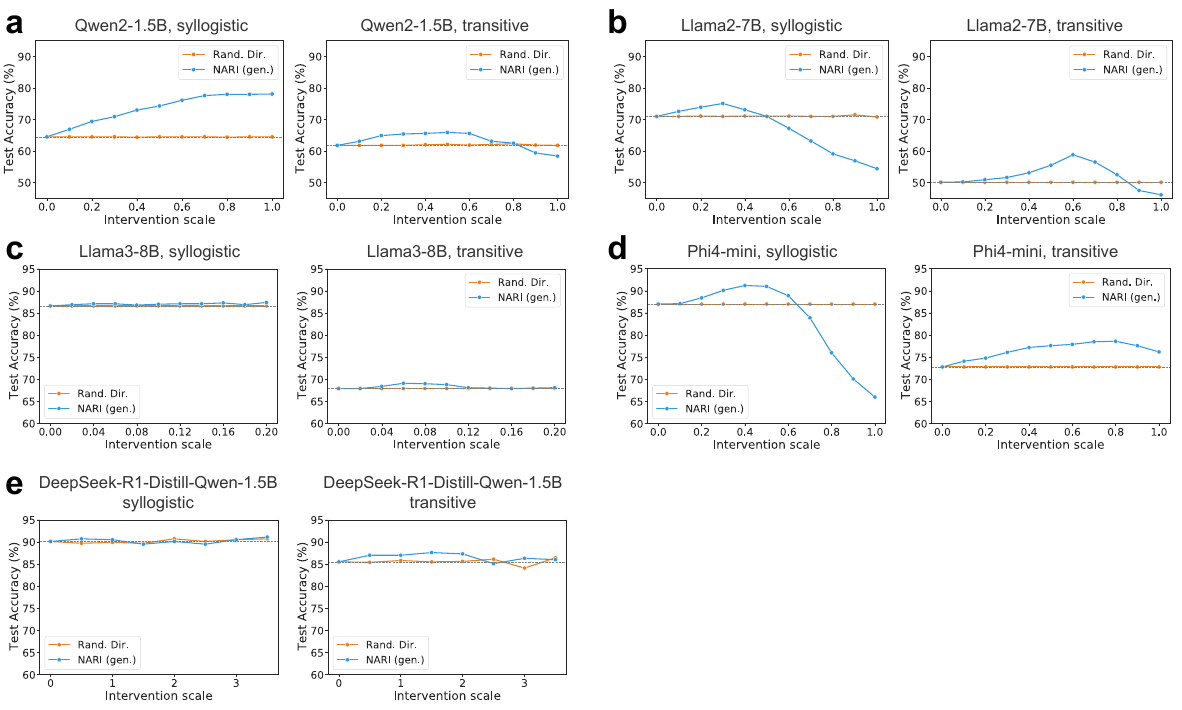}
    \caption[Analysis Results of General Direction Intervention Experiments]{\textbf{Analysis Results of General Direction Intervention Experiments.} Test accuracy of NARI (gen.) and random direction with different scales for various models.}
    \label{sup:fig:nari-alpha}
\end{figure}

In this section, we further provide analysis results of how test accuracy of NARI (gen.) scales as a function of the intervention scale $\alpha$. As shown in ~\cref{sup:fig:nari-alpha}, random directions produce little or no change in performance across $\alpha$, while brain-derived direction mostly has significant influence (typically, performance improves at moderate scales and then declines due to over intervention when $\alpha$ becomes too large). This suggests that brain-derived directions acts in a more reasoning-relevant subspace, which differs from random directions.

We also notice that the applicability of representation steering varies for different models, and the $\alpha$ parameter is sensitive and specific to models. Actually, it has been shown that the effectiveness of representation steering depends on the model and is not uniformly reliable~\cite{tan2024analysing,da2025steering}. It may be improved by more advanced steering methods in the future work.

\subsection{Analysis results of subject-wise robustness}

We further analyze the robustness of human subjects. We conduct experiments of NARI (gen.) with different number of subjects by varying the number of subjects from 1 to 10 through gradual inclusion of subjects with two directions: from high-performing to low-performing subjects and vice versa. This experiment also serves a similar purpose to the Leave-One-Subject-Out (LOSO) analysis, namely examining whether models overfit to specific subjects. As shown in ~\cref{sup:fig:subnum}a-b, increasing the number of subjects gradually improves the performance of NARI (gen.), where high-performing subjects generally have slightly better performance, while including low-performing subjects does not degrade the performance. These results verify the robustness of our method.

\section{Extended NARF results}\label{supp:sec:analysis-narf}

\subsection{More results on performance across reasoning subtypes}\label{supp:sec:narf-subtype}

\begin{figure}[t!]
    \centering
    \includegraphics[width=\linewidth]{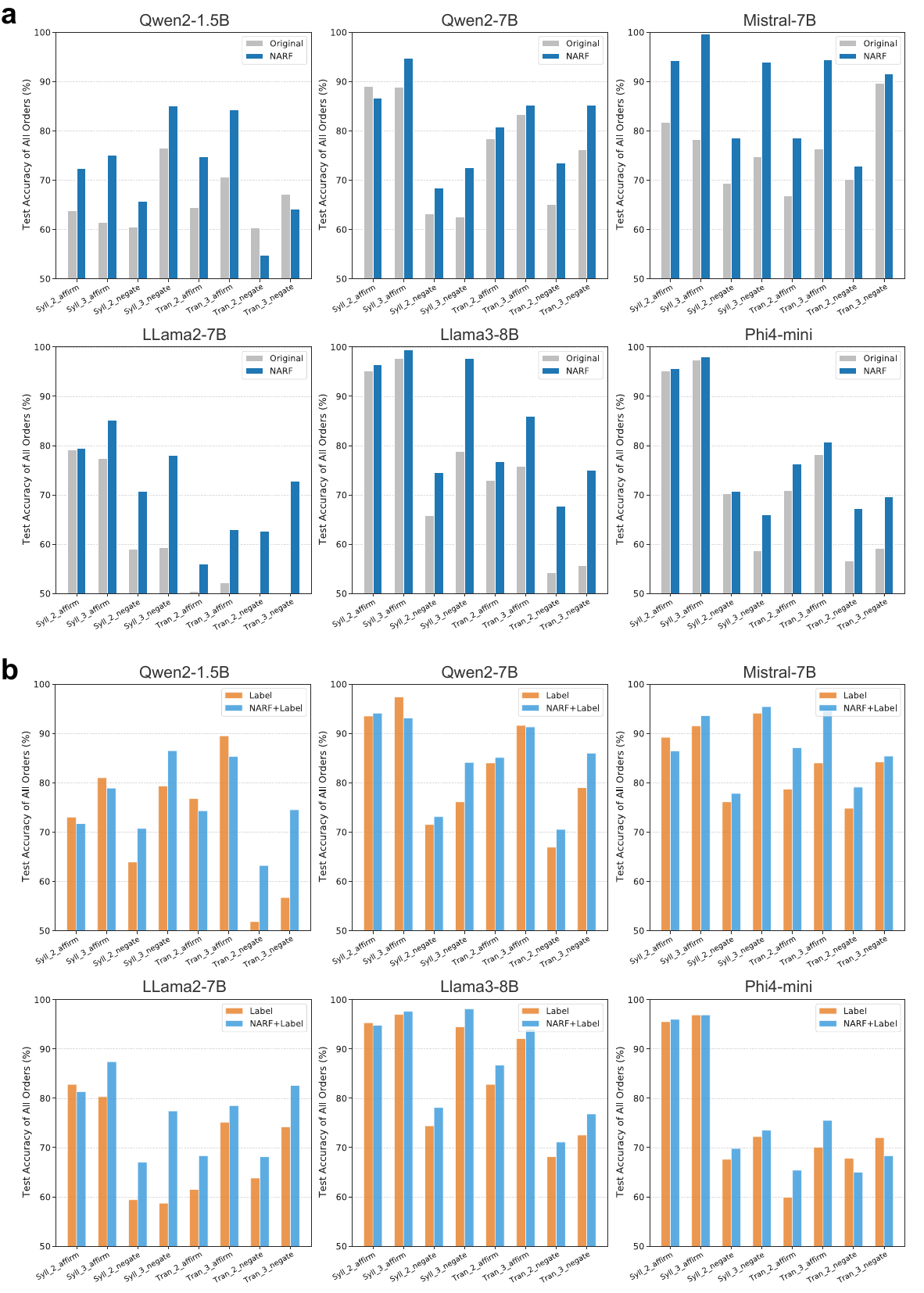}
    \caption[More Results on Performance across Reasoning Subtypes after Fine-tuning]{\textbf{More Results on Performance across Reasoning Subtypes after Fine-tuning.} (a) Comparison results between the original and \ac{narf} fine-tuned models. (b) Comparison results between the language-label-only fine-tuned and \ac{narf}+Label fine-tuned models.}
    \label{sup:fig:finetune-subtype}
\end{figure}

In \cref{sup:fig:finetune-subtype}, we provide more results of each model's performance across reasoning subtypes after fine-tuning. Subtypes are defined by the reasoning type and condition (Methods, \cref{supp:sec:generated-data}): for each reasoning type (syllogistic/transitive), conclusion form (affirmative/negation), and premise requirement (2/3 to make judgement), we gather the results of problems with ``true'' or ``false'' conclusions, leading to 8 subtypes such as Syll\_2\_affirm. \cref{sup:fig:finetune-subtype} shows that the improvement of \ac{narf} compared to original models or language-label-only fine-tuned models is consistent for most subtypes on the test data. To quantify this, we perform one-sided paired t-tests over different models ($n=6$) for NARF $>$ Original under each reasoning subtype, and the results show that NARF results are significantly higher than original results for almost all subtypes (Syll\_2\_affirm: $p=0.101$; Syll\_3\_affirm: $p=0.022$; Syll\_2\_negate: $p=0.004$; Syll\_3\_negate: $p=0.001$; Tran\_2\_affirm: $p=0.004$; Tran\_3\_affirm: $p=0.007$; Tran\_2\_negate: $p=0.033$; Tran\_3\_negate: $p=0.027$). For Syll\_2\_affirm, p-value is not significant because of the small sample size and small improvement under originally high accuracy for some models, while five of six models show improvement (up to 12.4\%). These results suggest that there is no subtype that NARF consistently fails to improve.

\subsection{Ablation analysis of NARF when combined with language supervision}\label{supp:sec:ablation-narflabel}

\begin{figure}[t!]
    \centering
    \includegraphics[width=\linewidth]{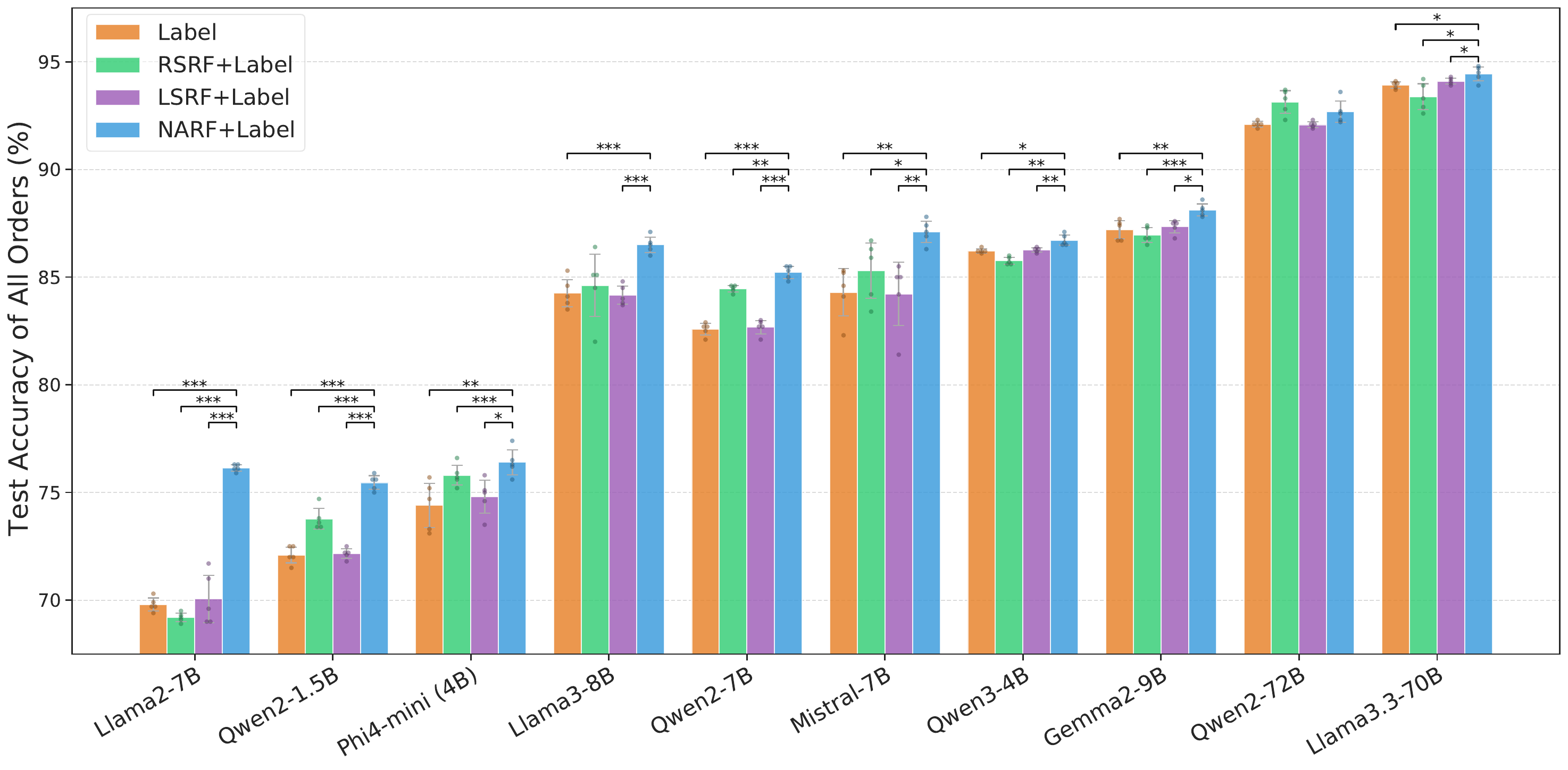}
    \caption[Ablation Analysis of \ac{narf} when Combined with Language Supervision]{\textbf{Ablation Analysis of \ac{narf} when Combined with Language Supervision.} Statistical tests are reported based on two-sided paired t-test over five runs of experiments under same random seeds. RSRF replaces \ac{fmri} signals with random data, which eliminates the effect of human neural activations and is only induced by the representational structures of models. LSRF replaces \ac{fmri} signals with one-hot encodings of label signals, representing the joint effect by the representational structures of models and labels. Results are ordered based on the models' original test accuracy.}
    \label{sup:fig:narflabel-ablation}
\end{figure}

As analyzed in \cref{sup:sec:explain-narinarf}, \ac{narf} is also induced by the joint effect of model and neural representational structures. To analyze the effect of the components, we perform ablation studies by replacing the human \ac{fmri} signals with random data (random signal induced representation fine-tuning (RSRF)) similar to the baseline in \ac{nari} experiments, or with one-hot encodings of label signals (label signal induced representation fine-tuning (LSRF)). RSRF eliminates the effect of human neural activations and is only induced by the model representational structure (\cref{sup:sec:explain-narinarf}). LSRF introduces label signals and is induced by the joint effect of model representational structure and labels. As shown in \cref{sup:fig:narflabel-ablation}, while RSRF+Label also improves label-only fine-tuning due to the information in model representations, \ac{narf} yields significantly larger gains, indicating that human neural structures provide unique improvement signals beyond model-internal patterns, and the improvement is also a joint effect of model and neural representations. Meanwhile, LSRF+Label does not show improvement over RSRF+Label (sometimes even has negative impacts on label fine-tuning), indicating that the structure of labels cannot effectively guide representations for generalization to new problems, highlighting the effectiveness of the structures of human neural activations.

\subsection{Similarity metrics with human neural activations}\label{supp:sec:similarity-metrics}

\begin{figure}[t!]
    \centering
    \includegraphics[width=\linewidth]{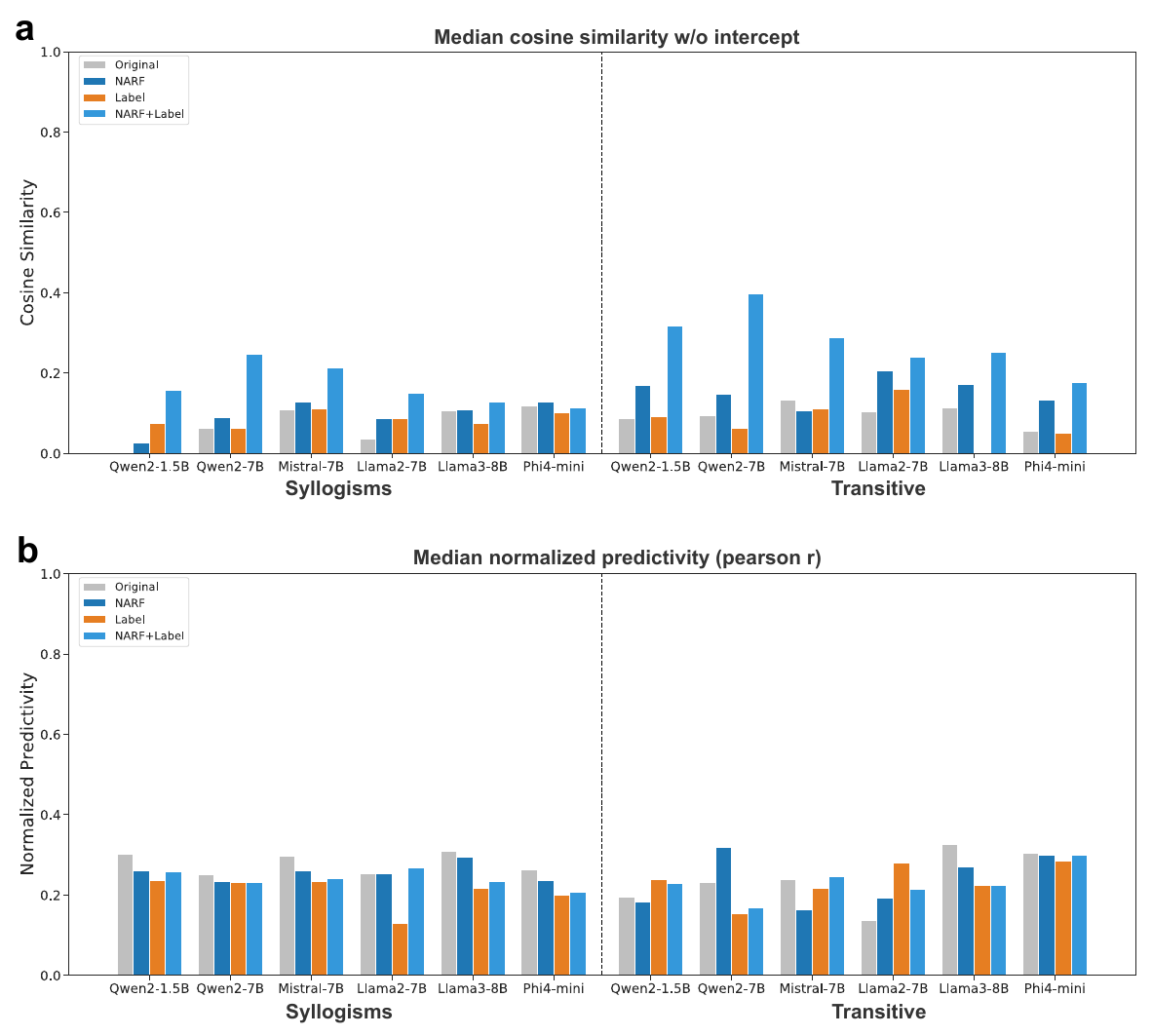}
    \caption[Results of Similarity Metrics with Human Neural Activations after Fine-tuning]{\textbf{Results of Similarity Metrics with Human Neural Activations after Fine-tuning.} (a) Results of the intercept-excluded similarity metric (our objective), which removes the intercept term and is based on cosine similarity. (b) Results of the similarity metric following the predictivity calculation, which includes the intercept term and is based on Pearson r correlation among data samples.}
    \label{sup:fig:finetune-similarity}
\end{figure}

As we optimize models' representations toward directions encouraging a higher similarity metric with human neural activations, we further analyze the similarity metrics after fine-tuning. We first analyze the similarity metric calculated as the objective in \ac{narf} (\ie, $S=\text{Sim}(\mathbf{W}\mathbf{x},\mathbf{y})$ that removes the intercept term) and we take the average of similarity among all considered layers. As shown in \cref{sup:fig:finetune-similarity}a, \ac{narf} and \ac{narf}+Label largely increase this similarity metric, while language-label-only training hardly realizes it, indicating that they are different directions. \ac{narf}+Label has much higher similarity because it directly optimizes this objective, while \ac{narf} only encourages representations to the nearest successfully intervened point induced by human neural activations. We also analyze the similarity metric following the common predictivity calculation (based on $\mathbf{W}\mathbf{x}+\mathbf{b}$ and the Pearson r correlation metric among data samples). As shown in \cref{sup:fig:finetune-similarity}b, \ac{narf} does not show consistent increase/decrease, indicating that the directions are different from optimizing this predictivity score.
Indeed, more accurate model does not necessarily have higher global alignment with human, because 1) humans are not perfect (no subject achieves 100\% accuracy; the average of human accuracy in the dataset is 86.4\%) and 2) functional guidance and global alignment metrics are not necessarily monotonically coupled. Our method applies neural guidance primarily to model's incorrect problems, whereas standard alignment score compute Pearson correlation over all samples. As a result, this score does not necessarily increase, while the sample-wise cosine similarity increases.

\subsection{More visualizations of reasoning trajectories}

\begin{figure}[t!]
    \centering
    \includegraphics[width=\linewidth]{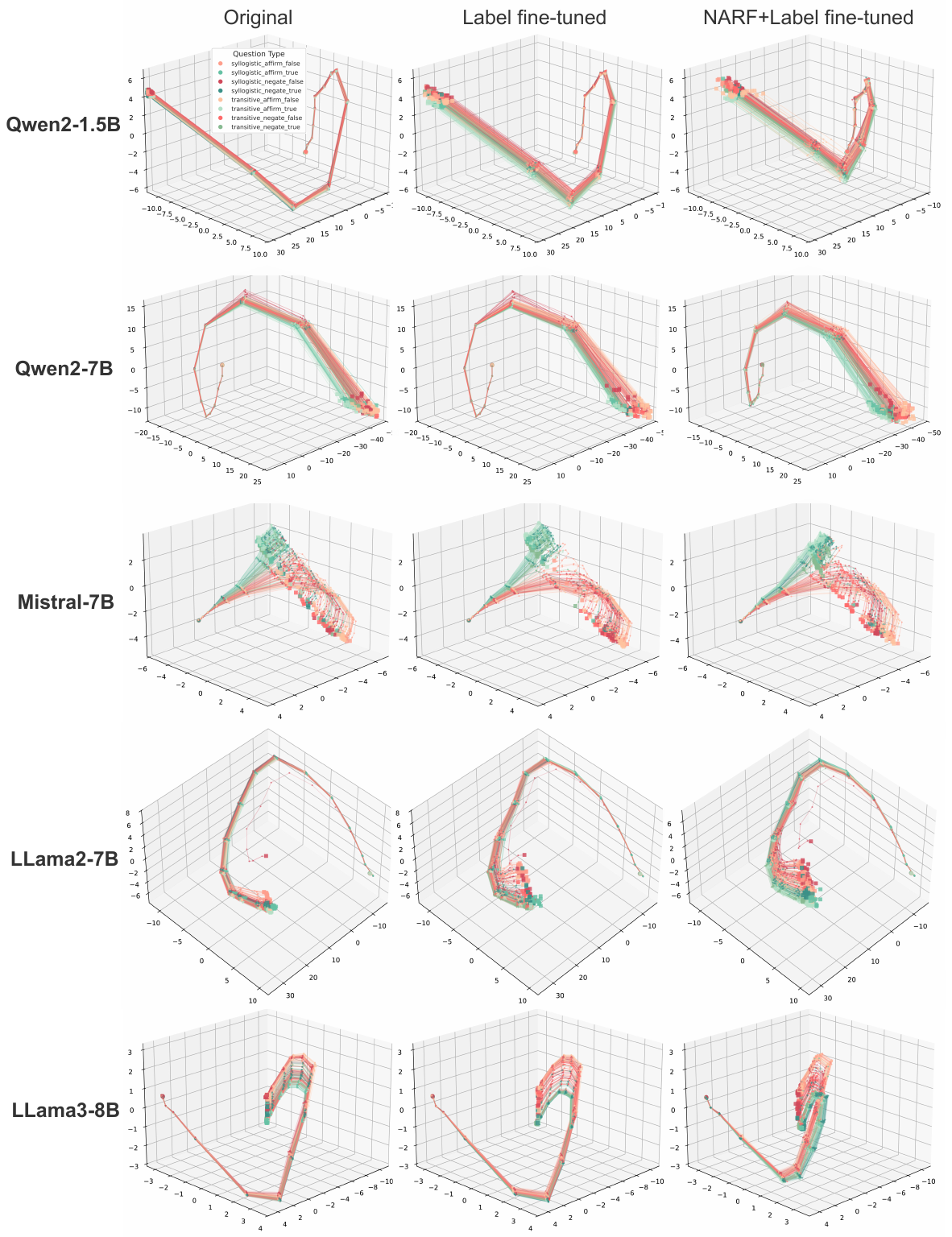}
    \caption[More Visualization Results of the Layer-wise Reasoning Trajectories in the Representation Space]{\textbf{More Visualization Results of the Layer-wise Reasoning Trajectories in the Representation Space.} The trajectories are layer outputs from the 1/4-th layer to the 3/4-th layer, projected into the three-dimensional space based on \ac{pca}.}
    \label{sup:fig:more-reasoning-path}
\end{figure}

We present more visualizations of the layer-wise reasoning trajectories for various models in \cref{sup:fig:more-reasoning-path}, demonstrating that compared with language-label-only fine-tuning, \ac{narf}+Label consistently improves the separation between paths for correct and incorrect logics, leading to higher performance.

\subsection{Evaluation of general capabilities}

To further evaluate whether fine-tuning with neural signals leads to catastrophic forgetting of general model capabilities, we evaluate all fine-tuned models on standard benchmarks, including MMLU, GSM8K, HumanEval, and MBPP, following the OpenCompass~\cite{2023opencompass} evaluation framework. The results (\cref{sup:tab:general-capability}) show that performance on these general benchmarks remains essentially stable after fine-tuning, with only small fluctuations, and in several cases, performance even slightly improves. These results indicate that our method does not suffer from catastrophic forgetting of general abilities. This stability is possibly due to both a general property of \acp{llm} that task-specific fine-tuning on limited data typically preserves broad capabilities, and the design of \ac{narf} in which the neural-guided signal enters the objective with directionally-bounded targets or constraints from language labels.

\subsection{Analysis results of subject-wise robustness}

\begin{figure}[t!]
    \centering
    \includegraphics[width=\linewidth]{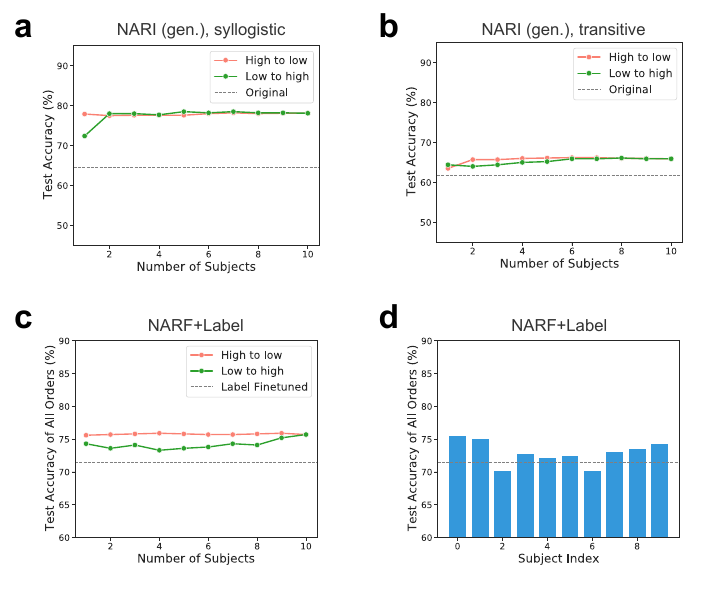}
    \caption[Analysis Results of the Influence of Different Human Subjects on NARI (gen.) and NARF+Label]{\textbf{Analysis Results of the Influence of Different Human Subjects on NARI (gen.) and NARF+Label.} (a-b) Test performance of NARI (gen.) on syllogistic and transitive problems with varying number of subjects in two directions: from high-performing to low-performing subjects and vice versa. (c) Test performance of NARF+Label with varying number of subjects in two directions: from high-performing to low-performing subjects and vice versa. (d) Test performance of NARF+Label for each subject.}
    \label{sup:fig:subnum}
\end{figure}

We also analyze the robustness of human subjects for NARF+Label. We vary the number of subjects from 1 to 10 through gradual inclusion of subjects with two directions: from high-performing to low-performing subjects and vice versa.  This experiment also serves a similar purpose to the LOSO analysis, namely examining whether models overfit to specific subjects. As shown in ~\cref{sup:fig:subnum}c, it is also similar for NARF+Label that including high-performing subjects leads to better performance while including low-performing subjects given high-performing subjects does not degrade the performance. ~\cref{sup:fig:subnum}d further demonstrates the performance for each subject, showing that the method is not only effective for high-performing subjects, and while not all subjects have uniform gains, including these subjects does not degrade model performance. These results verify the robustness of our method.

\subsection{Combination with synthetic data}\label{sup:sec:combine-synthetic}

We suggest neural signals as an additional source of information for representations that is complementary to common behavioural supervision, and the proposed method aims to collaborate and complement rather than compete with common machine learning techniques. To further evaluate the potential synergy of NARF and other techniques, we evaluate combining NARF+Label with additional synthetic data. Specifically, we generate synthetic data similar to the generation of test problems, while using different pseudowords and adjectives. The list of adjectives for syllogistic reasoning is: \textit{ancient, modern, fierce, gentle, elegant, clumsy, swift, lazy, curious, mysterious, fragile, sturdy, noisy, silent, graceful, awkward, clever, foolish, generous, selfish}, and the list of comparative adjectives for transitive reasoning is: \textit{easier, busier, noisier, safer, riskier, wiser, cleverer, luckier, healthier, cheaper, costlier, earlier, later, rarer, simpler, stricter, looser, fairer, nearer, further}. We also generate the same number of questions for all groups for a balanced data distribution. As the original fMRI dataset contains 70 problems, we generate 32, 64, 96, 128, 160, and 192 questions by gradually adding new instances to study the performance under different training data ratio over the original data. These questions only have language labels for supervision. We use the same training setting to train Mistral-7B for Label and NARF+Label under the same random seed, keeping the gradient accumulation step the same as 70. 

\begin{figure}[t!]
    \centering
    \includegraphics[width=0.5\linewidth]{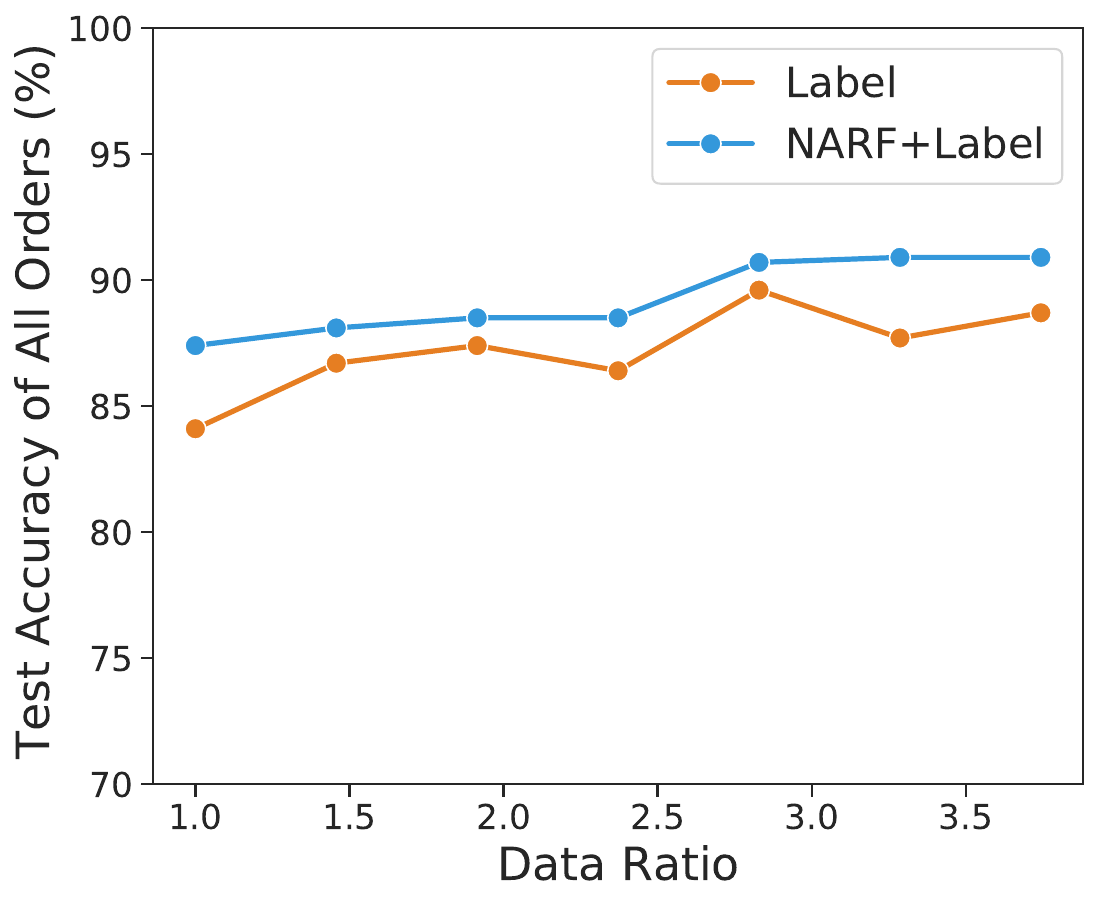}
    \caption[Results of Combining NARF with Synthetic Data]{\textbf{Results of Combining NARF with Synthetic Data.} Data ratio refers to the ratio of training data (problems from the fMRI dataset plus new synthetic data) over original problems from the fMRI dataset.}
    \label{sup:fig:synthetic}
\end{figure}

\cref{sup:fig:synthetic} presents the results of performance over data ratio, showing that neural guidance continues to provide gains over language-only supervision across data ratios, even with limited fMRI data. The results suggest that neural signals can provide orthogonal gains over common methods, with certain ``neural data efficiency''.

\section{Extended explanation of experimental logic}\label{sup:sec:extended-explanation}

\begin{figure}[t!]
    \centering
    \includegraphics[width=0.6\linewidth]{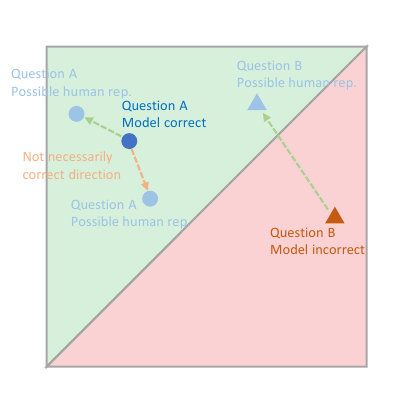}
    \caption[Illustration of Experimental Logic]{\textbf{Illustration of experimental logic.} We only intervene on problems that models are incorrect because for those that models are already correct, human representations do not necessarily provide correct direction.}
    \label{sup:fig:exp_demo}
\end{figure}

In this section, we further explain the logic of experiment design for only problems that the model is incorrect. 
The main reason is that humans are not perfect (e.g., no subject achieves 100\% accuracy) and more ``human-like'' representations are not guaranteed to point to more correct direction for already-correct answers. As shown in \cref{sup:fig:exp_demo} for a simplified illustration, suppose that we have a shared decision boundary between models and humans for correct and incorrect answers, for problems that both humans and models are correct, human representations can either be closer or farther to the boundary compared to model representations. Thus the direction to human representations can either strengthen the model decision or weaken the correct answer toward incorrect ones. Therefore, intervention on correct answers can in principle have both outcomes and cannot inversely suggest whether human and model representations have shared subspace.

By contrast, the model-incorrect / human-correct case is the most informative setting for testing the causal utility of neural guidance, as we can expect the direction toward human representations to guide incorrect points to correct ones if they have shared space. If steering toward the human-derived direction repairs a model error, this can inversely suggest a shared subspace. Additionally, for the other option, intervention on problems that model is correct while human is incorrect, it has trivial possibilities such as the language processing abilities of models are broken. Thus, intervention on incorrect answers toward correct ones is the only non-trivial case for our purpose.

For sanity check, we also apply a similar instance-specific NARI procedure to initially-correct problems with 50 iterations and different $\alpha$, and report the average flip-to-incorrect rate (mean across subjects). The results show moderate flip-to-incorrect rates (1.6\% for $\alpha=1$, 13.0\% for $\alpha=2$, 23.6\% for $\alpha=3$, 28.9\% for $\alpha=4$, 32.0\% for $\alpha=5$, 33.5\% for $\alpha=6$, 35.1\% for $\alpha=7$, 36.6\% for $\alpha=8$, 37.0\% for $\alpha=9$, and 37.6\% for $\alpha=10$), which is consistent with our explanation of experimental logic that intervention on correct answers can in principle have both outcomes. The flip rate is dependent on the relative position of model and human representations compared to the decision boundary, as well as the robustness of the model, and cannot suggest conclusions.

\section{Extension to the HCP relational processing dataset}\label{sup:sec:hcp}

\begin{figure}[t!]
    \centering
    \includegraphics[width=\linewidth]{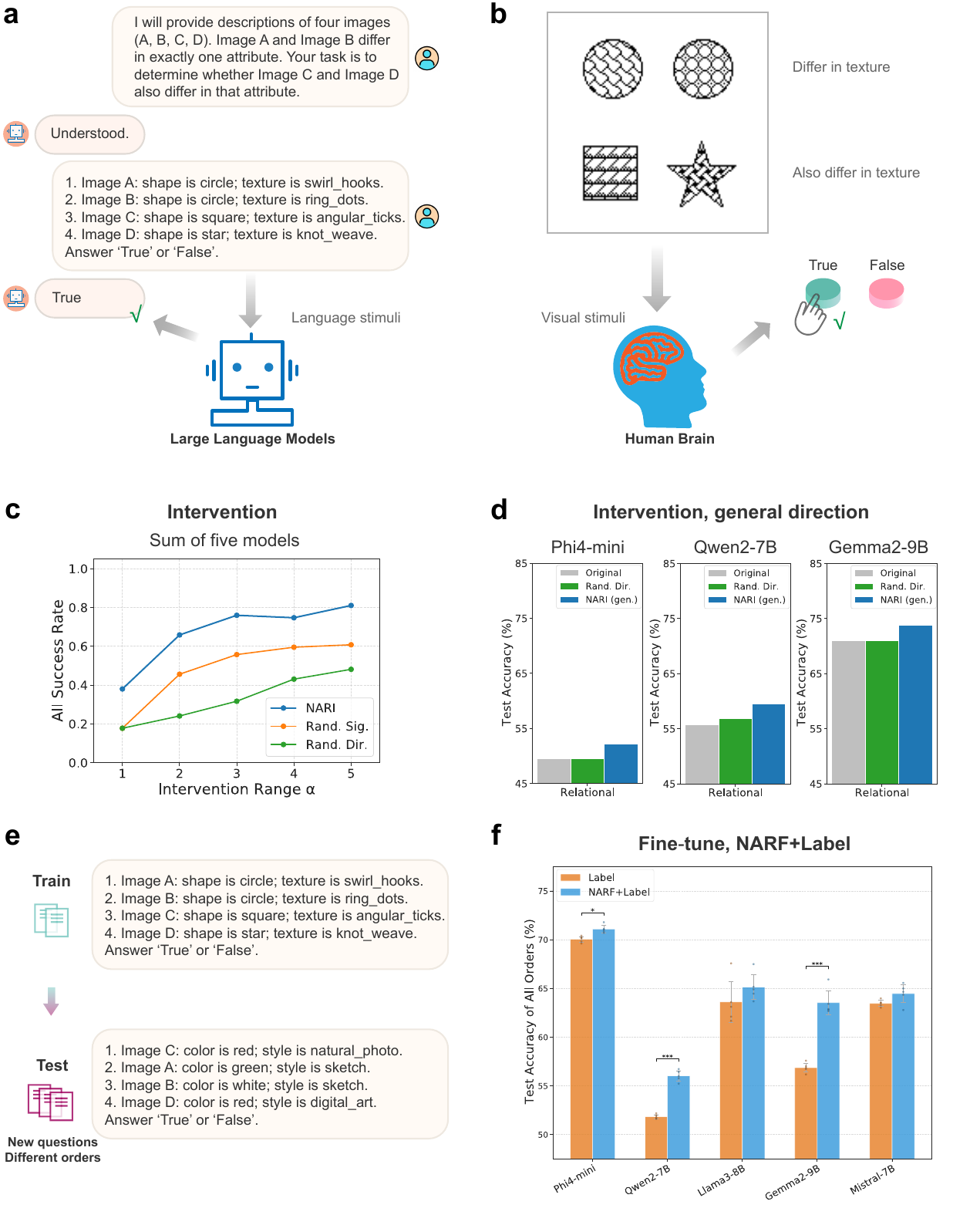}
    \caption[Results on the Relational Processing Task from the HCP Dataset]{\textbf{Results on the Relational Processing Task from the HCP Dataset.} (a-b) Illustration of the task for LLMs and humans. The task probes relational processing by comparing relations from two pairs of images, judging whether bottom images differ in the attribute that top image differ in. LLMs process problems through chat-formatted prompts with language-based descriptions of image attributes, generating direct linguistic responses. Human participants view two pairs of images (top and bottom), responding via button press. (c) Summary results of five LLM models for intervention on their incorrect problems. NARI achieves much higher success rates than baselines of random signals and the max-integration of random directions. (d) Results of the general direction on the new test data. (e) Illustration of the training and testing settings. (f) Testing accuracies (mean$\pm$std) of various models for language-label-only and NARF+Label fine-tuning. Statistical tests are reported based on two-sided paired t-test over five runs of experiments under same random seeds.}
    \label{sup:fig:hcp}
\end{figure}

In this section, to further validate our method on other independent datasets, we extend our experiments to HCP Relational Processing Task from the Human Connectome Project~\cite{van2013wu} (\cref{sup:fig:hcp}). The HCP Relational Processing Task probes relational reasoning by comparing relations from two pairs of images, judging whether bottom images differ in the attribute that top images differ in (\cref{sup:fig:hcp}b). We adapt it to large language models by presenting language descriptions of the attributes of the images to the models (\cref{sup:fig:hcp}a). Such different modalities extend our experimental settings to evaluate a different abstract relational processing ability with different stimulus modalities. This dataset differs from our original deductive reasoning dataset in both task and stimulus modality, thereby providing a meaningful external validation of the overall pipeline.

\paragraph{Details of the fMRI dataset and preprocessing}
The dataset is derived from the relational processing task in the Human Connectome Project (publicly available on \url{www.humanconnectome.org}), which originally contains more than 1,000 human subjects. We randomly select 20 subjects (IDs: 100610, 103111, 139637, 149236, 162733, 169444, 200008, 200513, 211619, 308129, 421226, 441939, 529549, 567759, 611938, 727553, 771354, 793465, 872158, 984472), whose behavioural accuracy ranges from 70\% to 100\% for this task. 
The dataset probes higher-order relational reasoning using visual stimuli that vary in shape and texture. It includes relational and matching (the control condition) blocks across two runs for each subject. For the relational block, each trial presents two pairs of images, where the top images differ in exactly one attribute (shape or texture). Participants are required to judge whether the bottom images also differ in that attribute and respond via button press. Each subject has a total of 24 relational trials (2 runs, 3 relational blocks each run, and 4 trials each block).

While the dataset follows a block-designed task-fMRI paradigm, we extract the stimuli for each trial from the accompanying description file of experimental designs for each subject (which describes the file name of each stimuli) and the publicly available stimuli files (from the HCP\_TFMRI\_scripts). As the stimuli files are images, we manually annotate the shape and texture attributes of each image and convert them into language descriptions. There are 96 stimuli files in the scripts, while we find that only 47 stimuli files appear in fMRI experiments of the 20 subjects. There are total six possible shapes and six possible textures for images.

We use the fMRI data preprocessed by the Human Connectome Project, and estimate the response to each trial with GLMSingle. We use the average response time as the response time in GLMSingle. For the design matrix, we consider up to 3 conditions (used for cross-validation in GLMSingle): correct answer for questions with ``yes'' answer, correct answer for questions with ``no'' answer, and wrong answer, and extract single-trial responses for each problem with the design matrix.

We extract voxels from the estimated response based on locations of MD regions and the subject-specific functional localization process by contrasting relational blocks over matching blocks.
We first use SPM12 to get the t-values of all voxels considering the contrast, and then select 10\% of the most responsive voxels within the MD mask. These voxels are considered as vectors of brain representations for each human subject.

\paragraph{Details of model prompting}
We convert image stimuli to language descriptions for LLMs to process. Specifically, we convert the six possible shapes into: \textit{circle, cross, triangle, square, star, hexagon}, and convert the six possible textures into: \textit{ring\_dots, angular\_ticks, zigzag\_dashes, solid\_blocks, knot\_weave, swirl\_hooks}. The task prompt for the model is:

``\textit{You are an expert in relational reasoning. I will provide descriptions of four images (A, B, C, D), which describe two attributes of each image: shape and texture. Image A and Image B differ in exactly one attribute (either shape or texture). Your task is to determine whether Image C and Image D also differ in that attribute. For example, if Image A and Image B have different shapes, then you should determine if Image C and Image D also have different shapes. Or if Image A and Image B have different textures, then you should determine if Image C and Image D also have different textures. If the statement is true, you should answer `True'; otherwise answer `False'. Only answer `True' or `False' at the beginning of the response.}''

After the model's round saying ``\textit{Understood. I will analyze the attributes and perform relational reasoning to judge the statement. I will only answer `True' or `False' at the beginning of the response.}'', we present the descriptions as follows:

``\textit{1. Image A: shape is circle, texture is solid\_blocks.$\backslash$n 2. Image B: shape is circle, texture is ring\_dots.$\backslash$n 3. Image C: shape is triangle, texture is zigzag\_dashes.$\backslash$n 4. Image D: shape is hexagon, texture is zigzag\_dashes.$\backslash$n For the attribute that Image A and Image B differ in (either shape or texture), do Image C and Image D also differ in that attribute? Only answer `True' or `False' at the beginning of your response.}''

Then we extract the first token of the model's response.

\paragraph{Generated test dataset}
For testing of LLMs, we generate new questions with different descriptions of attribute types and attributes. Specifically, we replace ``shape'' and ``texture'' by two random attribute type in \textit{[color, quality, geometry, style]}, and each of them has the following possible attributes: \textit{color: [red, green, blue, yellow, black, white]; quality: [normal, noisy, blurred, underexposed, overexposed, compressed]; geometry: [rotated, translated, zoomed\_in, zoomed\_out, flipped, sheared; style: [sketch, watercolor, oil\_painting, digital\_art, cartoon, natural\_photo]}. There are four conditions: true/false $\times$ first/second attribute is different. For the test data, we generate 100 problems for each condition, leading to 400 problems; for the validation data for fine-tuning, we generate 20 problems for each condition, leading to 80 problems. When evaluating fine-tuned models, we also enumerate all possible permutations for the order of descriptions of four images (\cref{sup:fig:hcp}e), with finally $400\times 24=9600$ test data.

\paragraph{Implementation details of NARI and NARF}
We leverage the similar experimental settings as the deductive reasoning task. For NARI, the maximum gradient descent steps are set as 50. For the general direction of NARI, we take the average of successful intervention directions with $\alpha=3$ (for Phi4-mini, we take $\alpha=5$), normalize the direction for each layer to the standard deviation scale, and the scale $\gamma$ is searched between 0.5 and 10 with the interval 0.5. For the combination of NARF and language labels, we use all 96 questions from the HCP stimuli files with only part of them having fMRI signals. We select 10 subjects for fine-tuning based on the intervention success rate, and only perform guidance for questions that the model can be intervened to the correct answer. We start training of NARF+Label from the language-label-trained models, except for Phi4-mini for which we find that starting from the original model has better performance. $\lambda_{narf}$ is set as 0.1 for Phi4-mini and Llama3-8B, 0.05 for Qwen2-7B, and 0.01 for Mistral-7B and Gemma2-9B. We also run each experiment for 5 times with the same random seeds: 0, 1, 42, 1000, 2024, and report statistical significance based on two-sided paired t-test among the 5 runs.

\paragraph{Results}
We conduct intervention and fine-tuning experiments on this task and dataset using five models from different families (Phi4-mini, Qwen2-7B, Llama3-8B, Gemma2-9B, Mistral-7B). As shown in \cref{sup:fig:hcp}c, human neural activations from this dataset can also induce effective representation intervention directions, significantly outperforming both baselines (model representational structure alone / random directions) across perturbation ranges. While the overall intervention success rate on this dataset (around 80\%) is lower than the original deductive reasoning setting (100\%), this gap is likely due to the increased difficulty introduced by the modality shift, reflecting a boundary condition under which our method remains effective. In addition, \cref{sup:fig:hcp}d further shows that the general direction can transfer to new, out-of-set problems. Finally, in a similar fine-tuning setting with evaluation on new questions under different orders (\cref{sup:fig:hcp}e), NARF complements language supervision with consistent gains over language supervision alone (\cref{sup:fig:hcp}f).

These results on an independent dataset show that our pipeline is not limited to one specific experimental design. Meanwhile, they show that our method can extend to other tasks with different stimulus modalities, suggesting a promising way to incorporate more potential neural signal data.

\clearpage

\clearpage

\clearpage

\clearpage

\clearpage

\clearpage

\clearpage

\clearpage

\clearpage

\clearpage

\clearpage

\begin{table}[t]
	\centering
	\caption[Evaluation of General Capabilities of Various Models after Fine-tuning]{\textbf{Evaluation of General Capabilities of Various Models after Fine-tuning.}}
    \label{sup:tab:general-capability}
	\begin{tabular}{llcccc}
	\toprule
	\multirow{2}{*}{Model} & \multirow{2}{*}{Method} & \multicolumn{4}{c}{Task Performance} \\
	\cmidrule(lr){3-6}
	 & & MMLU & GSM8K & HumanEval & MBPP \\
	\midrule
	\multirow{4}{*}{Llama2-7B} & Original & 36.84 & 29.19 & 14.02 & 19.8 \\
	 & NARF & 37.18 & 29.04 & 14.02 & 19.2 \\
	 & Label & 41.08 & 28.13 & 14.63 & 20.2 \\
	 & NARF+Label & 37.39 & 28.43 & 13.41 & 19.2 \\
	\midrule
	\multirow{4}{*}{Qwen2-1.5B} & Original & 55.54 & 61.11 & 43.90 & 30.8 \\
	 & NARF & 55.65 & 61.41 & 42.68 & 30.6 \\
	 & Label & 55.62 & 61.64 & 44.51 & 30.6 \\
	 & NARF+Label & 55.76 & 62.77 & 45.73 & 31.2 \\
	\midrule
	\multirow{4}{*}{Phi4-mini} & Original & 74.47 & 83.02 & 78.05 & 55.4 \\
	 & NARF & 74.16 & 82.71 & 78.66 & 54.6 \\
	 & Label & 74.50 & 83.24 & 78.05 & 55.0 \\
	 & NARF+Label & 74.56 & 83.02 & 78.05 & 55.0 \\
	\midrule
	\multirow{4}{*}{Llama3-8B} & Original & 67.60 & 79.23 & 58.54 & 58.0 \\
	 & NARF & 67.71 & 78.17 & 56.71 & 57.4 \\
	 & Label & 67.82 & 78.24 & 60.30 & 57.4 \\
	 & NARF+Label & 67.80 & 78.10 & 58.54 & 56.6 \\
	\midrule
	\multirow{4}{*}{Qwen2-7B} & Original & 70.48 & 80.97 & 78.66 & 53.2 \\
	 & NARF & 70.36 & 80.36 & 76.83 & 53.2 \\
	 & Label & 70.44 & 81.12 & 78.66 & 53.0 \\
	 & NARF+Label & 70.33 & 79.83 & 77.44 & 55.4 \\
	\midrule
	\multirow{4}{*}{Mistral-7B} & Original & 59.11 & 45.41 & 30.49 & 22.8 \\
	 & NARF & 59.36 & 45.34 & 34.76 & 25.2 \\
	 & Label & 59.46 & 46.47 & 30.49 & 23.6 \\
	 & NARF+Label & 59.46 & 46.10 & 31.10 & 23.4 \\
	\midrule
	\multirow{3}{*}{Qwen3-4B} & Original & 75.79 & 86.81 & 79.27 & 65.4 \\
	 & Label & 76.06 & 87.19 & 79.27 & 65.4 \\
	 & NARF+Label & 75.84 & 86.43 & 78.05 & 65.6 \\
	\midrule
	\multirow{3}{*}{Gemma2-9B} & Original & 74.74 & 79.23 & 23.78 & 59.8 \\
	 & Label & 74.25 & 79.45 & 25.61 & 59.2 \\
	 & NARF+Label & 74.33 & 80.29 & 25.61 & 59.4 \\
	\midrule
	\multirow{3}{*}{Qwen2-72B} & Original & 82.34 & 91.66 & 83.54 & 63.6 \\
	 & Label & 82.50 & 91.28 & 82.32 & 67.6 \\
	 & NARF+Label & 82.29 & 91.96 & 82.93 & 65.8 \\
	\midrule
	\multirow{3}{*}{Llama3.3-70B} & Original & 81.36 & 92.87 & 85.37 & 76.4 \\
	 & Label & 81.50 & 92.80 & 84.76 & 76.2 \\
	 & NARF+Label & 81.73 & 92.72 & 84.15 & 77.0 \\
	\bottomrule
	\end{tabular}
\end{table}

\clearpage

\clearpage

\clearpage

\clearpage

\clearpage
{
    \bibliographystyle{unsrt}
    \bibliography{reference_header,reference}
}

\end{document}